%% file: ecsqaru21_mlcchain.tex
\documentclass[runningheads]{llncs}
\usepackage{graphicx}
\usepackage[colorlinks = true,
            linkcolor = blue,
            urlcolor  = blue,
            citecolor = blue,
            anchorcolor = blue]{hyperref}

\usepackage{scalerel}
\usepackage{mathtools}
\usepackage{amssymb}
\usepackage{amsmath}
\usepackage{comment}
\usepackage{tikz}
\usepackage{arydshln}
\usepackage{xparse}
\usepackage{mathrsfs}
\usepackage{stmaryrd}
\usepackage{bm}
\usepackage{bbm}
\usepackage{float}
\usepackage{subfigure}
\usepackage{tikz}
\usepackage{pgfplots}
\usetikzlibrary{automata,chains}
\usetikzlibrary{shapes.misc}
\usepackage{nicefrac}
\usepackage{hyperref}

\newcommand{\tred}[1]{\textcolor{red}{#1}}

\newcommand{\loss}{\ell}

\newcommand{\inputspace}{\mathscr{X}}
\newcommand{\outspace}{\mathcal{K}}
\DeclareMathAlphabet{\mathpzc}{OT1}{pzc}{m}{it}
\newcommand{\newinstance}{\vect{x}}
\newcommand{\cat}{\mathpzc{m}}

\newcommand{\argmin}{\operatorname*{arg\,min~}}
\newcommand{\argmax}{\operatorname*{arg\,max~}}

\newcommand{\prob}{\mathbb{P}}
\newcommand{\condprob}[1][]{
	\ifthenelse{\equal{#1}{}}{P_{\newinstance}}{P_{#1}}
}
\newcommand{\hatcondprob}[1][]{
	\ifthenelse{\equal{#1}{}}{\hat{P}_{\newinstance}}{\hat{P}_{#1}}
}
\newcommand{\expe}{{\mathbb{E}}}

\newcommand{\credal}{{\mathscr{P}}}

\NewDocumentCommand{\lowerexp}{O{\credal}}{\underline{\mathbb{E}}_{#1}}
\NewDocumentCommand{\upperexp}{O{\credal}}{\overline{\mathbb{E}}_{#1}}

\newcommand{\eq}{\!=\!}
\newcommand{\setn}[1]{\llbracket #1 \rrbracket}
\newcommand{\vect}[1]{{\boldsymbol  #1}}

\newcommand{\reals}{\mathbb{R}}

\newcommand{\spacelabel}{\mathscr{Y}}
\newcommand{\spacepartial}{\mathfrak{Y}}
\newcommand{\setindices}{\mathscr{I}}

\usepackage{xparse}

\NewDocumentCommand{ \condprobchain }{ O{P} O{j\!-\!1}}{%
 #1_{\newinstance}^{\setn{#2}\vspace*{1mm}}
}

\newcommand{\condprobchainhat}[1][]{
	\ifthenelse{\equal{#1}{}}{
		\condprobchain[\hat{P}]
	}{
		\condprobchain[\hat{P}][#1]~~
	}
}
\newcommand{\lowercondprobchain}[1][]{
	\ifthenelse{\equal{#1}{}}{
		\condprobchain[\underline{P}]
	}{
		\condprobchain[\underline{P}][#1]~~
	}
}
\newcommand{\uppercondprobchain}[1][]{
	\ifthenelse{\equal{#1}{}}{
		\condprobchain[\overline{P}]
	}{
		\condprobchain[\overline{P}][#1]~~
	}
}
\newcommand{\subidx}[1]{\scaleto{\setindices_{\mathcal{#1}}^{j-1}}{7pt}}

\begin{document}
\title{Multi-label Chaining with Imprecise Probabilities}
%
%
\author{Yonatan Carlos {Carranza Alarc\'on}\inst{1}\orcidID{0000-0002-8657-6355} \and
S\'ebastien Destercke\inst{1}\orcidID{0000-0003-2026-468X}}
%
%
\institute{Sorbonne Universit\'es, Universit\'e Technologique de Compi\`egne, CNRS, UMR 7253 - Heudiasyc, 57 Avenue de Landshut, Compi\`egne, France
\email{\{yonatan-carlos.carranza-alarcon, sebastien.destercke\}@hds.utc.fr}}
\maketitle              
\begin{abstract}
We present two different strategies to extend the classical multi-label chaining approach to handle imprecise probability estimates. These estimates use convex sets of distributions (or credal sets) in order to describe our uncertainty rather than a precise one. The main reasons one could have for using such estimations are (1) to make cautious predictions (or no decision at all) when a high uncertainty is detected in the chaining and (2) to make better precise predictions by avoiding biases caused in early decisions in the chaining. We adapt both strategies to the case of the naive credal classifier, showing that this adaptations are computationally efficient. Our experimental results on missing labels, which investigate how reliable these predictions are in both approaches, indicate that our approaches produce relevant cautiousness on those hard-to-predict instances where the precise models fail.
\end{abstract}

Multi-label classification (MLC) is a generalization of traditional classification (with a single label), as well as a special case of multi-task learning. This approach is increasingly required in different research fields, such as the classification of proteins in bioinformatics~\cite{tsoumakas2007multi}, text classification in information retrieval~\cite{furnkranz2008multilabel}, object
recognition in computer vision~\cite{boutell2004learning}, and so on.

A classical issue in multi-label learning techniques is how to integrate the possible dependencies between labels while keeping the inference task tractable. Indeed, while decomposition techniques~\cite{tsoumakas2007multi,furnkranz2008multilabel} such as Binary relevance or Calibrated ranking allow to speed up both the learning and inference tasks, they roughly ignore the label dependencies, while using a fully specified model such as in probabilistic trees~\cite{dembczynski2010bayes} requires, at worst, to scan all possible predictions (whose quantity grows exponentially in the number of labels). A popular technique, known as chaining~\cite{read2011classifier} to solve this issue, at least for the inference task, is to use heuristic predictions: they consists in using, incrementally, the predictions made on previous labels as additional conditional features to help better predict the relevance of a current label, rather than using the prediction probabilities. 

To the best of our knowledge, there are only a few works on multi-label classification producing cautious predictions, such as the reject option~\cite{pillai2013multi}, partial predictions~\cite{destercke2014multilabel,antonucci2017multilabel} or abstaining labels~\cite{nguyen2019reliable}, but none of these have studied this issue in chaining (or classifier-chains approach).

In this paper, we consider the problem of extending chaining to the imprecise probabilistic case, and propose two different extensions, as some predictive probabilities are too imprecise to use predicted labels, henceforth called abstained labels, in the chaining. The first extension treats the abstained labels in a robust way, exploring all possible conditional situation in order not to propagate early uncertain decisions, whereas the latter marginalizes the probabilistic model over those labels, ignoring them in the predictive model. 

Section \ref{sec:preliminareschain} introduces the notations that we use for the multi-label setting, and gives the necessary reminders about making inferences with convex sets of probabilities. In Section \ref{sec:multilabelchaining}, we recall the classical classifier-chains approach and then we present our extended approaches based on imprecise probabilities. Section~\ref{sec:mlcncc} then shows that in the case of the naive credal classifier (NCC)~\cite{zaffalon2002naive}, those strategies can be performed in polynomial time. 

Finally, in Section \ref{sec:experimentschain}, we perform a set of experiments on real data sets, which are perturbed with missing labels, in order to investigate how cautious (when we abstain on labels difficult to predict) is our approach. In order to adhere to the page limit, all proofs and supplementary experimental results have been relegated to the appendix of an online extended version~\cite{alarcon2021multilabel}.



\section{Problem Setting}\label{sec:preliminareschain}

In the multi-label problem, an instance $\vect{x}$ of an input space $\inputspace\eq\mathbb{R}^p$ is no longer associated with a single label $\cat_k$ of an output space $\outspace\eq\{\cat_1, \dots, \cat_m\}$, as in the traditional classification problem, but with a subset of labels $\Lambda_x\!\subseteq\!\mathcal{K}$ often called the set of relevant labels while its complement $\mathcal{K} \backslash \Lambda_x$ is considered as irrelevant for $\vect{x}$. Let $\spacelabel\eq\{0, 1\}^m$  be a $m$-dimensional binary space and $\boldsymbol y \eq (y_1, \dots, y_m) \in \spacelabel$ be any element of $\spacelabel$ such that  $y_i\eq1$ if and only if $\cat_i \in \Lambda_x$.

From a decision theoretic approach (DTA), the goal of the multi-label problem is the same as usual classification. Given a probability distribution $\hat{\mathbb{P}}$ fitting a set of i.i.d. observations $\mathcal{D}=\{(\vect{x}_i, \vect{y}_i)|i=1,\dots,N\}$ issued from a (true) theoretical probability distribution $\mathbb{P}: \inputspace \times \spacelabel \rightarrow [0, 1]$, DTA aims to minimize the risk of getting missclassification with respect to a specified loss function $\loss(\cdot, \cdot)$:
\begin{equation}\label{eq:classrisk}
\setlength{\abovedisplayskip}{6pt}\setlength{\abovedisplayshortskip}{6pt}
\setlength{\belowdisplayskip}{6pt}\setlength{\belowdisplayshortskip}{6pt}
\mathcal{R}_{\loss}(Y, h(X)) = \underset{\boldsymbol h}{\min} ~ \mathbb{E}_{\hat{\mathbb{P}}} \left[\loss(Y, \bm h(X)) \right].
\end{equation}
where $\bm h\!:\!\inputspace\!\rightarrow\!\spacelabel$ is a m-dimensional vector function. If $\loss(\cdot, \cdot)$ is defined instance-wise, that is $\loss:\mathcal{Y} \times \mathcal{Y} \to \reals$, the solution of  Equation~\eqref{eq:classrisk} is obtained by minimizing the conditional expected risk (cf. \cite[eq. 3]{dembczynski2012label} and \cite[eq. 2.21]{friedman2001elements})
\begin{equation} \label{eq:minconditional}
\setlength{\abovedisplayskip}{6pt}
\setlength{\abovedisplayshortskip}{6pt}
\setlength{\belowdisplayskip}{6pt}
\setlength{\belowdisplayshortskip}{6pt}
\boldsymbol h(\newinstance) = \underset{\vect{y} \in \spacelabel}{\arg \min} ~ \mathbb{E}_{\hat{\mathbb{P}}_{Y|\newinstance}} \left[\loss(Y, \vect{y}) \right] = \arg\min_{\vect{y} \in \spacelabel} \sum_{\vect{y}' \in \spacelabel} \hat{P}(Y\!=\!\vect{y}'|X\!=\!\newinstance) \loss(\vect{y}',\vect{y})
\end{equation}
or, equivalently, by picking the undominated elements of the order relation\footnote{A complete, transitive, and asymmetric binary relation} $\succeq$ over $\spacelabel\!\times\!\spacelabel$ for which $\vect{y}^1\!\succeq\!\vect{y}^2$ ($\vect{y}^1$ is preferred to/dominates $\vect{y}^2$)  iff
\begin{equation}
\setlength{\abovedisplayskip}{6pt}\setlength{\abovedisplayshortskip}{6pt}
\setlength{\belowdisplayskip}{6pt}\setlength{\belowdisplayshortskip}{6pt}
	\expe_{\hat{\mathbb{P}}_{Y|\newinstance}} \left(\loss(\vect{y}^2,\cdot) - \loss(\vect{y}^1,\cdot) \right) = \expe_{\hat{\mathbb{P}}_{Y|\newinstance}}\left(\loss(\vect{y}^2,\cdot)\right) - \expe_{\hat{\mathbb{P}}_{Y|\newinstance}} \left(\loss(\vect{y}^1,\cdot) \right) \geq 0.
\end{equation}
This amounts to saying that exchanging $\vect{y}^2$ for $\vect{y}^1$ would incur a non-negative expected loss (which is not desirable).

In this paper, we are also interested in making set-valued predictions when uncertainty is too high (e.g. due to insufficient evidence to include or discard a label as relevant, see Example~\ref{exa:notevidence}). The set-valued prediction will here be described as a partial binary vector $\vect{y}^*\!\in\spacepartial$ where $\spacepartial\eq\{0, 1, *\}^m$ is the new output space with a new element $*$ representing the abstention. For instance, a partial prediction $\vect{y}^*\eq(*, 1, 0)$ corresponds to two plausible binary vector solutions $\{(0, 1, 0), (1, 1, 0)\}\!\subseteq\!\spacelabel$. To obtain such predictions, we will use imprecise probabilities as a well-founded framework.



\subsection{Notions about Imprecise Probabilities}
Imprecise probabilities consist in representing our uncertainty by a convex set of probability distributions $\mathscr{P}_{X}$ \cite{walley1991statistical,augustin2014introduction} (i.e. a \emph{credal set} \cite{levi1983enterprise}), defined over a space $\inputspace$ rather than by a precise probability measure $\mathbb{P}_{X}$ \cite{taylor1973introduction}. Given such a set of distributions  $\mathscr{P}_{X}$ and any measurable event $A \subseteq \inputspace$, we can define the notions of lower and upper probabilities as:
\begin{equation}
\setlength{\abovedisplayskip}{6pt}\setlength{\abovedisplayshortskip}{6pt}
\setlength{\belowdisplayskip}{6pt}\setlength{\belowdisplayshortskip}{6pt}
\underline{P}_{X}(A) = \inf_{P \in \mathscr{P}_{X}} P(A)  \quad \quad \text{and} \quad \quad \overline{P}_{X}(A) = \sup_{P \in \mathscr{P}_{X}} P(A)
\end{equation}
where $\underline{P}_{X}(A) = \overline{P}_{X}(A)$ only when we have sufficient information about event $A$. The lower probability is dual to the upper~\cite{augustin2014introduction}, in the sense that $\underline{P}_{X}(A) = 1 -\overline{P}_{X}(A^c)$ where $A^c$ is the complement of $A$. Many authors \cite{walley1991statistical,zaffalon2002naive} have argued that when information is lacking or imprecise, considering credal sets as our model of information better describes our actual uncertainty. 

However, such an approach comes with extra challenges in the learning and inference step, especially in combinatorial domains. In this paper, we will consider making a chain of binary inferences, each inference influenced by the previous one. If we consider $\mathcal{Y}\eq\{0, 1\}$ as the output space and $Y$ as a univariate random variable on $\mathcal{Y}$, a standard way to take a decision with abstention given a credal set $\credal$ on $\mathcal{Y}$ is 
\begin{equation}\label{eq:hammingstrong}
	\setlength{\abovedisplayskip}{6pt}
	\setlength{\abovedisplayshortskip}{6pt}
	\setlength{\belowdisplayskip}{6pt}
	\setlength{\belowdisplayshortskip}{6pt}
	\hat{y} = \begin{cases}
		~~~1 & \text{ if }
		\underline{P}_{\newinstance}(Y\!=\!1) > 0.5,\\
		~~~0 & \text{ if }
		\overline{P}_{\newinstance}(Y\!=\!1) < 0.5,\\
		~~~* & \text{ if } 0.5\in
		\left[\underline{P}_{\newinstance}(Y\!=\!1),
		 \overline{P}_{\newinstance}(Y\!=\!1)\right]
\end{cases}.
\end{equation}
The next example illustrates such notions on a multi-label example
\begin{example}\label{exa:notevidence}
We consider an output space of two labels $\outspace= \{\cat_1, \cat_2\}$, a single binary feature $x_1$ and Table \ref{tbl:imprecise} with imprecise estimations of $\hat{\prob}(Y_1, Y_2|X_1)$.
\begin{table}[!ht]
\vspace{-1mm}\centering\setlength{\tabcolsep}{5pt}
\begin{tabular}{cc:c|c||cc:c|c}
\hline $y_1$ & $y_2$ & $x_1$ & $\hat{\credal}_{\scaleto{Y_1, Y_2|X_1\eq0}{5pt}}$
& $ y_1$ & $y_2$ & $x_1$ &$\hat{\credal}_{\scaleto{Y_1, Y_2|X_1\eq1}{5pt}}$   \\ \hline
0  & 0 & 0 & [0.4,0.7]  & 0  & 0 & 1 & 0.00 \\
0  & 1 & 0 & [0.3,0.6]  & 0  & 1 & 1 & 0.00 \\
1  & 0 & 0 & 0.00 & 1  & 0 & 1 & [0.6,0.8] \\
1  & 1 & 0 & 0.00 & 1  & 1 & 1 & [0.2,0.4] \\\hline
\end{tabular}\vspace{2mm}
\caption{Estimated conditional probability distributions $\hat{\prob}(Y_1, Y_2|X_1)\in\hat{\credal}_{Y_1, Y_2|X_1} $.}
\label{tbl:imprecise}\vspace{-8mm}
\end{table}

Based on the probabilities of Table \ref{tbl:imprecise}, we have that $\hat{P}_0(y_1\!=\!0):=\hat{P}(y_1\!=\!0|x_1\!=\!0)\eq1$ and $\hat{P}_0(y_2\!=\!0)\! \in [0.4,0.7]$, therefore not knowing whether $\hat{P}_0(y_2\!=\!0)>0.5$. This could lead us to propose as a prediction $\hat{\vect{y}}^*\eq(0,*)$. On the contrary, the imprecision on the right hand-side is such that {$\hat{P}_1(y_2\!=\!0)\! \in [0.6,0.8]$}, leading to the precise prediction $\hat{\vect{y}}^*\eq(1,0)$. 
\end{example}

\section{Multilabel Chaining with Imprecise Probabilities}
\label{sec:multilabelchaining}
Solving~\eqref{eq:minconditional} can already be computationally prohibitive in the precise case~\cite{dembczynski2010bayes}, which is why heuristic to approximate inferences done on the full joint models such as the chain model have been proposed. This section recalls its basics and presents our proposed extension. To do so, we will need a couple notations: we will denote by $\setindices$ subsets of label indices and by $\setn{j}=\{1, \dots, j\}$ the set of the first $j$ integers. Given a prediction made in the $j$ first labels, we will denote by
\begin{enumerate}
	\item (relevant labels) $\setindices_\mathcal{R}^j \subseteq \setn{j}$ the indices of the labels predicted as relevant among the $j$ first, i.e. $\forall i\in\setindices_{\mathcal{R}}^j,~y_i=1$,
	\item (irrelevant labels)  $\setindices_\mathcal{I}^j \subseteq \setn{j}, \setindices_\mathcal{I}^j\cap\setindices_\mathcal{R}^j=\emptyset~$ the indices of the labels predicted as irrelevant among the $j$ first, i.e. $\forall i\in\setindices_{\mathcal{I}}^j,~y_i=0$, and
	\item (abstained labels) $\setindices_\mathcal{A}^j = \setn{j} \backslash (\setindices_\mathcal{R}^j \cup \setindices_\mathcal{I}^j)$  the indices of the labels on which we abstained among the $j$ first, i.e. $\forall i\in\setindices_{\mathcal{A}}^j,~y_i=\{0, 1\} := *$, 
\end{enumerate}
and of course $\setindices^j \eq \setindices_\mathcal{A}^j \cup \setindices_\mathcal{R}^j \cup \setindices_\mathcal{I}^j \eq \setn{j}$. Besides, for the sake of simplicity, we will use the notation
\begin{equation}
\setlength{\abovedisplayskip}{6pt}\setlength{\abovedisplayshortskip}{6pt}
\setlength{\belowdisplayskip}{6pt}\setlength{\belowdisplayshortskip}{6pt}
\condprobchain(Y_j\!=\!1):=P(Y_j\!=\!1|Y_{\setindices^{j-1}}\!\!=\hat{\bm{y}}_{\setindices^{j-1}}, X=\bm x),
\end{equation}
where $\hat{\bm{y}}_{\setindices^{j-1}}$ is a $(j-1)$-dimensional vector that contains the previously inferred precise and/or abstained values of labels having indices $\setindices^{j-1}$.
\subsection{Precise Probabilistic Chaining}
\begin{figure*}[!ht]
	\vspace{-4mm}
	\centering
	\resizebox{1.02\textwidth}{!}{
	\hspace{-8mm}
	\subfigure[Chaining with $\{Y_1, Y_2\}$]{
		\input{images/precise_chaining}
		\label{fig:precisechain}
	}\hspace{-4mm}%
	\subfigure[Chaining with $\{Y_2, Y_1\}$]{
		\input{images/precise_chaining_inverse}
		\label{fig:inverselabels}
	}}\vspace{-4mm}
	\caption{Precise chaining}\label{fig:precisechaining}\vspace{-4mm}
\end{figure*}
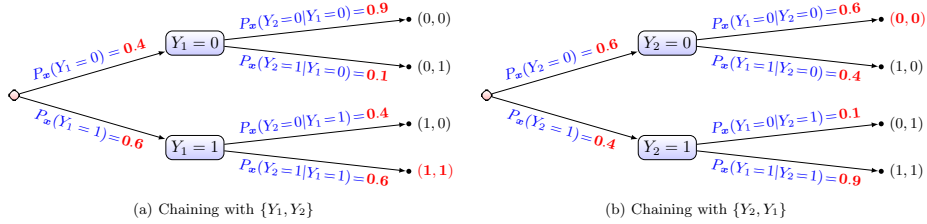

Classifier chains is a well-known approach exploiting dependencies among labels by fitting at each step of the chain (see Figure \ref{fig:precisechaining}) a new classifier model $h_j:\inputspace\times\{0,1\}^{j-1} \rightarrow\{0,1\}$ predicting the relevance of the j$th$ label. This classifier combines the original input space attribute and all previous predictions in the chain in order to create a new input space $\inputspace_{j-1}^*=\inputspace\times\{0,1\}^{j-1}, j\in\mathbb{N}^{>0}$. In brief, we consider a chain $\bm h=(h_1, \dots, h_m)$ of binary classifiers resulting in the full prediction $\hat{\bm y}$ obtained by solving each single classifier as follows
\begin{equation}
\setlength{\abovedisplayskip}{4pt}\setlength{\abovedisplayshortskip}{4pt}
\setlength{\belowdisplayskip}{4pt}\setlength{\belowdisplayshortskip}{4pt}
\hat{y}_{j} := h_j(x) = \underset{y \in \{0,1\}}{\arg \max}~
	\condprobchain(Y_{j}\!=\!y).
\end{equation}

The classical multi-label chaining then works as follows:
\begin{enumerate}
	\item {\sc Random label ordering.} We randomly pick an order between labels 
	and assume that the indices are relabelled in an increasing order.  
	\item {\sc Prediction $j^{th}$ label.} For a label $y_j$ and the previous predictions on labels $y_1, \dots, y_{j-1}\!$ and let  $\setindices_\mathcal{R}^{j-1}, \setindices_{\mathcal{I}}^{j\!-\!1}\!\subseteq\!\llbracket j\!-\!1\rrbracket$ be set of indices of relevant and irrelevant labels with $\setindices_\mathcal{R}^{j-1}\!\cap\!\setindices_\mathcal{I}^{j-1}\!=\!\emptyset$. Then, the prediction of $\hat{y}_j$ (or $h_j(\newinstance)$) for a new instance $\newinstance$ is
\begin{equation}
	\setlength{\abovedisplayskip}{6pt}\setlength{\abovedisplayshortskip}{6pt}
	\setlength{\belowdisplayskip}{6pt}\setlength{\belowdisplayshortskip}{6pt}
	\hat{y}_j =
	\begin{cases}
	1 &\text{if}~ P_{\newinstance}(Y_j=1|
		Y_{\setindices_\mathcal{R}^{j-1}}=1,
		Y_{\setindices_{\mathcal{I}}^{j-1}} =0)=\condprobchain(Y_j\!=\!1)
	\geq 0.5 \\
	0 &\text{if}~ P_{\newinstance}(Y_j=1|
		Y_{\setindices_\mathcal{R}^{j-1}}=1,
		Y_{\setindices_{\mathcal{I}}^{j-1}} =0)=\condprobchain(Y_j\!=\!1)
	< 0.5
 	\end{cases}
\end{equation}
\end{enumerate}

Figure~\ref{fig:precisechaining} summarizes the procedure presented above, as well as the obtained predictions for a specific case (in bold red predicted labels and probabilities).

Figure~\ref{fig:precisechaining} shows that using this heuristic can lead to strong biases, as two different orderings of the same joint model can lead to shift from one prediction to its opposite. Intuitively, adding some robustness and cautiousness in the process could help to avoid unwarranted biases. 

In what follows, we propose two different extensions of precise chaining based on  imprecise probability estimates, in which the final prediction belongs to the output space $\spacepartial$ of partial bianry vectors. 

\subsection{Imprecise Probabilistic Chaining}
We now consider that the estimates $\condprobchain(Y_j\!=\!1)$ can become imprecise, that is, we now have $[\condprobchain](Y_j\!=\!y_j)\!:=\![\lowercondprobchain(Y_j\!=\!y_j),$ $\uppercondprobchain(Y_j\!=\!y_j)]$. The basic idea of using such estimates is that in the chaining, we should be cautious when the classifier is unsure about the most probable prediction. In this section, we describe two different strategies (or extensions) in a general way, and we will propose an adaptation of those strategies to the NCC in the next section.

Let us first formulate the generic procedure to calculate the probability bound of the $j^{th}$ label,
\begin{enumerate}
	\item {\sc Random label ordering.} As in the precise case. 
	\item {\sc Prediction $j^{th}$ label.}
	For a given label $y_j$, we assume we have (possily imprecise) predictions for $y_1,\dots, y_{j-1}$ such that $\setindices_{\mathcal{A}}^{j-1}$ are the indices of labels on which we abstained $\{*\}$ so far, and $\setindices_{\mathcal{R}}^{j-1}$ and $\setindices_{\mathcal{I}}^{j-1}$ remain the indices of relevant and irrelevant labels, such that $\setindices_{\mathcal{A}}^{j-1}\cup\setindices_{\mathcal{R}}^{j-1}\cup\setindices_{\mathcal{I}}^{j-1}\eq\setindices^{j-1}$. Then, we calculate $[\condprobchain](Y_j\!=\!1)$  in order to predict the label $\hat{y}_j$ as
\begin{equation}\label{eq:hammcondit}
\setlength{\abovedisplayskip}{4pt}
\setlength{\abovedisplayshortskip}{4pt}
\setlength{\belowdisplayskip}{4pt}
\setlength{\belowdisplayshortskip}{4pt}
\hat{y}_j = \begin{cases}
	~~~1 & \text{if }
	\lowercondprobchain(Y_j=1) > 0.5,\\
	~~~0 & \text{if }
	\uppercondprobchain(Y_j=1) < 0.5,\\
	~~~* & \text{if } 0.5\in
	[\lowercondprobchain(Y_j=1), 
		\uppercondprobchain(Y_j=1)],
\end{cases}
\end{equation}
where this last equation is a slight variation of Equation~\eqref{eq:hammingstrong} by using the new input space $\inputspace_{j-1}^*$.
\end{enumerate}

We propose the following two different extensions of how to calculate $[\condprobchain](Y_j=1)$ at each inference step of the imprecise chaining.
\subsubsection{Imprecise Branching}
The first strategy treats unsure predictions in a robust way, considering all possible branchings in the chaining as soon as there is an abstained label. Thus, the estimation of $[\lowercondprobchain(Y_j=1), \uppercondprobchain(Y_j=1)]$ (for $Y_j=0$, it directly obtains as $\lowercondprobchain(Y_j=0)=1-\uppercondprobchain(Y_j=1)$, and similarly for the upper bound) comes down to computing
\begin{equation}\label{eq:impbranching}
\setlength{\abovedisplayskip}{4pt}
\setlength{\abovedisplayshortskip}{4pt}
\setlength{\belowdisplayskip}{4pt}
\setlength{\belowdisplayshortskip}{4pt}
\begin{aligned}
	\lowercondprobchain(Y_j=1)&\!=\!\!\!\!
		\min_{\scaleto{\vect{y}\in\{0,1\}^{\left|\setindices_{\mathcal{A}}^{j-1}\right|}}{10pt}}
		\underline{P}_{\newinstance}(Y_j=1|
			Y_{\setindices_\mathcal{R}^{j-1}}=1,
			Y_{\setindices_{\mathcal{I}}^{j-1}}=0,
			Y_{\setindices_{\mathcal{A}}^{j-1}}=\vect{y}),\\
	\uppercondprobchain(Y_j=1) &\!=\!\!\!\!
	\max_{\scaleto{\vect{y}\in\{0,1\}^{\left|\setindices_{\mathcal{A}}^{j-1}\right|}}{10pt}}
		\overline{P}_{\newinstance}(Y_j=1|
				Y_{\setindices_\mathcal{R}^{j-1}}=1,
				Y_{\setindices_{\mathcal{I}}^{j-1}}=0,
				Y_{\setindices_{\mathcal{A}}^{j-1}}=\vect{y}).
\end{aligned}\tag{IB}
\end{equation}
So we consider all possible replacements of variables for which we have abstained so far. This corresponds to a very robust version of the chaining, where every possible path is explored. It will propagate imprecision along the tree, and may produce quite imprecise evaluations, especially if we abstain on the first labels.

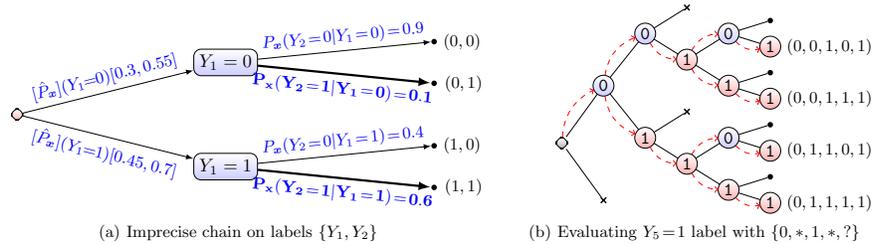
\begin{figure}[!th]
	\vspace{-4mm}\centering
	\resizebox{\textwidth}{!}{
	\subfigure[Imprecise chain on labels $\{Y_1, Y_2\}$]{
		\input{images/mix_precise_imprecise_chaining}
		\label{fig:mixchaining}
	}%
	\subfigure[Evaluating $Y_5\!=\!1$ label
	  with $\{0, *, 1, *, ?\}$]{
		\input{images/branching_chaining}
		\label{fig:pathimprecise}
	}}\vspace{-4mm}
	\caption{Imprecise branching strategy}\vspace{-4mm}
\end{figure}

Illustrations providing some intuition about this strategy can be seen in Figure~\ref{fig:pathimprecise} where we have abstained on labels $(Y_2,Y_4)$ and we want to compute lower and upper probability bounds of the label $Y_5=1$.

In Figure~\ref{fig:mixchaining}, we will consider the previous example (see Figure~\ref{fig:precisechaining}) in order to study in detail how we should calculate probability bounds $[\lowercondprobchain(Y_j=1), \uppercondprobchain(Y_j=1)]$. For the sake of simplicity, we assume that probabilities about $Y_2$ are precise and that the probability bounds for $Y_1=1$ is $\condprobchainhat[0](Y_1\!=\!1)\!\in\![0.45, 0.70]$. Taking all possible paths (bold on Figure~\ref{fig:mixchaining}) for which $Y_2=1$, we get
\begin{equation*}
	\setlength{\abovedisplayskip}{4pt}
	\setlength{\abovedisplayshortskip}{4pt}
	\setlength{\belowdisplayskip}{2pt}
	\setlength{\belowdisplayshortskip}{2pt}
	\begin{aligned}
		\lowercondprobchain[1](Y_2=1) &=
		\min_{y_1\in\{0,1\}} P_{\newinstance}(Y_2=1|Y_1=y_1) =
		\min(0.1, 0.6) = 0.1,\\[-1mm]
		\uppercondprobchain[1](Y_2=1) &=
		\max_{y_1\in\{0,1\}} P_{\newinstance}(Y_2=1|Y_1=y_1) =
		\max(0.1, 0.6) = 0.6,
	\end{aligned}	
\end{equation*}
which means that in this case we would abstain on both labels, i.e. $(\hat{y}_1, \hat{y}_2)\!\eq\!(*,*)$.
\subsubsection{Marginalization}
The second strategy simply ignores unsure predictions in the chaining. Its interest is that it will not propagate imprecision in the tree. 
Thus, we begin by presenting the general formulation (which will after lead to the formulation without unsureness) which takes into account unsure predicted labels conditionally, so the estimation of probability bounds $[\lowercondprobchain(Y_j=1), \uppercondprobchain(Y_j=1)]$ comes down to computing
\begin{equation*}\label{eq:marginalization}
\setlength{\abovedisplayskip}{5pt}
\setlength{\abovedisplayshortskip}{5pt}
\setlength{\belowdisplayskip}{5pt}
\setlength{\belowdisplayshortskip}{5pt}	
	\begin{aligned}
	\lowercondprobchain(Y_j\!=\!1)\!=\!
		\underline{P}_{\newinstance}(Y_j=1|
			Y_{\setindices_\mathcal{R}^{j-1}}\!=\!1,
			Y_{\setindices_{\mathcal{I}}^{j-1}}\!=\!0,
			Y_{\setindices_{\mathcal{A}}^{j-1}}\!=\!\{0,1\}^
				{|\setindices_{\mathcal{A}}^{j-1}|}),\\
	\uppercondprobchain(Y_j\!=\!1)\!=\!
		\overline{P}_{\newinstance}(Y_j=1|
			Y_{\setindices_\mathcal{R}^{j-1}}\!=\!1,
			Y_{\setindices_{\mathcal{I}}^{j-1}}\!=\!0,
			Y_{\setindices_{\mathcal{A}}^{j-1}}\!=\!\{0,1\}^
				{|\setindices_{\mathcal{A}}^{j-1}|}),
	\end{aligned}\!\!\tag{MAR}
\end{equation*} 
where $\setindices_{\mathcal{A}}^{j-1}=\{i_1, \dots, i_k\}$ denotes the set of indices of abstained labels and the last conditional term of probability bounds can be defined as
\begin{equation}
\setlength{\abovedisplayskip}{6pt}\setlength{\abovedisplayshortskip}{6pt}
\setlength{\belowdisplayskip}{6pt}\setlength{\belowdisplayshortskip}{6pt}	
\left(Y_{\setindices_{\mathcal{A}}^{j-1}}=\{0,1\}^{|\setindices_{\mathcal{A}}^{j-1}|}\right)\!:=\!(Y_{i_1}=0 \cup Y_{i_1}=1)\cap\dots\cap(Y_{i_k}=0 \cup Y_{i_k}=1).\!\!\!
\end{equation}

The \ref{eq:marginalization} formulation can be reduced by using Bayes's theorem in conjunction with the law of total probability. That is, for instance, given abstained labels $(Y_1=*, Y_3=*)$ and the precise prediction $(Y_2=1)$, inferring $Y_4=1$ comes down to computing $P_{\newinstance}(Y_4\eq1|(Y_1\eq0\cup Y_1\eq1),Y_2\eq1, (Y_3\eq0\cup Y_3\eq1))$ as follows
\begin{equation*}\small
\setlength{\abovedisplayskip}{6pt}
\setlength{\abovedisplayshortskip}{6pt}
\setlength{\belowdisplayskip}{6pt}
\setlength{\belowdisplayshortskip}{6pt}	
\frac{\sum\limits_{y_3,y_1\in\{0,1\}^2}\!\!\!\!\!\!\!\!
	P_{\newinstance}(Y_4\eq1,Y_1\eq y_1,Y_2\eq 1,Y_3\eq y_3)}
{\sum\limits_{y_3,y_1\in\{0,1\}^2}\!\!\!\!\!\!\!
	P_{\newinstance}(Y_1\eq y_1,Y_2\eq1, Y_3\eq y_3)}	
	\eq\frac{P_{\newinstance}(Y_4\eq1,Y_2\eq1)}
	{P_{\newinstance}(Y_2\eq1)}
	\eq P_{\newinstance}(Y_4\eq1|Y_2\eq1),
\end{equation*}
An illustration providing some intuition about this last example can be seen in Fig.~\ref{fig:margfiga}, in which we draw the possible path to infer the label $Y_4$ (considering a third branch in the chain to represent abstained labels).

The results of the last example can easily be generalized, and hence, \ref{eq:marginalization} comes down to calculating the new formulation called {\tred{(MAR*)}\label{eq:marginalization2}}
\begin{align}
	\lowercondprobchain(Y_j=1) &=
		\min\nolimits_{P\in\credal^*} P _{\newinstance}(Y_j=1|
			Y_{\setindices_\mathcal{R}^{j-1}}=1,
			Y_{\setindices_{\mathcal{I}}^{j-1}}=0),
		\label{eq:marlower2}\\[-1mm]
	\uppercondprobchain(Y_j=1) &=
		\max\nolimits_{P\in\credal^*} P_{\newinstance}(Y_j=1|
			Y_{\setindices_\mathcal{R}^{j-1}}=1,
			Y_{\setindices_{\mathcal{I}}^{j-1}}=0).
		\label{eq:marupper2}
\end{align}
where $\credal^*$ is simply the set of joint probability distributions described by the imprecise probabilistic tree (we refer to \cite{de2008imprecise} for a detailed analysis). \eqref{eq:marginalization} comes down to restrict the conditional chain model to consider only those labels on which we have not abstained (for which we have sufficient reliable information). In general, both~\eqref{eq:impbranching} and~\eqref{eq:marginalization} can lead to burdensome computations. In the next section, we propose to adapt their principle to the NCC, showing that this can be done efficiently. 

\begin{remark}
Note that another strategy that would be computationally efficient would be to simply considers all precise chaining paths consistent with local intervals, pruning dominated branches. However, this would also mean that each explored branch would have all previous predicted labels as conditioning features, thus still being impacted by the bias of picking those predictions. 
\end{remark}

 

\section{Imprecise Chaining with NCC} \label{sec:mlcncc}

NCC extends the classical naive Bayes classifier (NBC) on a set of distributions. NCC preserves the assumption of feature independence made by NBC, and relies on the Imprecise Dirichlet model (IDM)~\cite{walley1996inferences} to estimate class-conditional imprecise probabilities, whose imprecision level is controlled through a hyper-parameter $s\in\mathbb{R}$.
Therefore, the class-conditional probability bounds evaluated for  $Y_{j}=1$ ($Y_{j}=0$ can be directly calculated using duality) can be calculated as follows\footnote{For reviewer convenience, details are given in the supplementary material.}
{\par\nobreak\small\noindent\begin{align}
	\!\!\!\!\underline{P}(Y_{j}\eq1|
		\mathbf{X}\eq\newinstance, 
		\!Y_{\setindices^{j-1}}\!\eq\hat{\bm{y}}_{\setindices^{j-1}}\!) 
		&\eq
		\left(\!1\!+\!\frac{P(Y_{j}\eq0)
				\overline{P}_{0}(\mathbf{X}\eq{\newinstance})
				\overline{P}_{0}(Y_{\setindices^{j-1}}\!\eq
					\hat{\bm{y}}_{\setindices^{j-1}})
			}
			{P(Y_{j}\eq1)
				\underline{P}_{1}(\mathbf{X}\eq{\newinstance})
				\underline{P}_{1}(Y_{\setindices^{j-1}}\!\eq
					\hat{\bm{y}}_{\setindices^{j-1}})
			}
		\!\right)^{-1}\!\!\!\!\!\!\!\!,\!\!\label{eq:ncclower} \\
	\!\!\!\!\overline{P}(Y_{j}\eq1| 
			\mathbf{X}\eq\newinstance, 
		    \!Y_{\setindices^{j-1}}\!\eq
			\hat{\bm{y}}_{\setindices^{j-1}}\!) &\eq
		\left(\!1\!+\!\frac{P(Y_{j}=0)
			\underline{P}_{0}(\mathbf{X}\eq{\newinstance})
				\underline{P}_{0}(Y_{\setindices^{j-1}}\!\eq
					\hat{\bm{y}}_{\setindices^{j-1}})
			}
			{P(Y_{j}\eq1)
				\overline{P}_{1}(\mathbf{X}={\newinstance})
					\overline{P}_{1}(Y_{\setindices^{j-1}}\!\eq
						\hat{\bm{y}}_{\setindices^{j-1}})
			}
		\!\right)^{-1}\!\!\!\!\!\!\!\!.\!\!\label{eq:nccupper}
\end{align}}%
where conditional upper probabilities of $[\underline{P}_{1}, \overline{P}_{1}]$ and $[\underline{P}_{0}, \overline{P}_{0}]$ are defined as 
\begin{equation}\small
\setlength{\abovedisplayskip}{4pt}
\setlength{\abovedisplayshortskip}{4pt}
\setlength{\belowdisplayskip}{4pt}
\setlength{\belowdisplayshortskip}{4pt}	
\!\overline{P}_{a}(\mathbf{X}\eq\newinstance)\!:=\!
	\!\prod_{i=1}^p \overline{P}(X_i\eq x_i|Y_{j}\eq a)
\text{~and~}
\overline{P}_{a}(\mathbf{Y}_{
	 {\textstyle\mathstrut}\!\!\setindices^{j-1}}
	  \!\!\eq
	\vect{y}_{{\textstyle\mathstrut}\setindices^{j-1}}\!)\!:=\!\!
	\prod_{k=1}^{j-1} \overline{P}(Y_k\eq\hat{y}_k|Y_{j}\eq a),
\!\!\!\!\!\label{eq:ncclabelupper}
\end{equation}
where $a\in\{0, 1\}$. Conditional lower probabilities are obtained similarly. Using Equations~\eqref{eq:ncclower} and \eqref{eq:nccupper}, we now propose efficient procedures to solve the aforementioned strategies.


\subsection{Imprecise Branching}
In the specific case where we use the NCC, we can efficiently reduce the optimization problems of Equations~\eqref{eq:impbranching}, as expressed in the proposition below.
\begin{proposition}\label{prop:nccib}
Optimisation problems of the imprecise branching~\eqref{eq:impbranching} can be reduced by using probability bounds obtained from the NCC, namely Equations~\eqref{eq:ncclower} and \eqref{eq:nccupper}, as follows
\begin{equation*}
	\setlength{\abovedisplayskip}{6pt}
	\setlength{\abovedisplayshortskip}{6pt}
	\setlength{\belowdisplayskip}{6pt}
	\setlength{\belowdisplayshortskip}{6pt}	
    \begin{aligned}
    	\lowercondprobchain(Y_j\eq1)
    	\!\propto\!\!\!\!\!\!\!\!
    	\max_{\scaleto{\vect{y}\in\{0,1\}^{\left|\setindices_{\mathcal{A}}^{j-1}\right|}}{10pt}} \! 
			\frac{\overline{P}_{0}(
				Y_{\setindices_{\mathcal{A}}^{j-1}}\!\eq\vect{y})}
				{\underline{P}_{1}(
				Y_{\setindices_{\mathcal{A}}^{j-1}}\!\eq\vect{y})}
    	\text{~and~~}
    	\uppercondprobchain(Y_j\eq1) 
    	\!\propto\!\!\!\!\!\!\!\!
    	\min_{\scaleto{\vect{y}\in\{0,1\}^{\left|\setindices_{\mathcal{A}}^{j-1}\right|}}{10pt}}\!			 
    		\frac{\underline{P}_{0}(
			Y_{\setindices_{\mathcal{A}}^{j-1}}\!\eq\vect{y})}
			{\overline{P}_{1}(
			Y_{\setindices_{\mathcal{A}}^{j-1}}\!\eq\vect{y})}.
    \end{aligned}
\end{equation*}
Besides, applying the equations 
derived from the imprecise Dirichlet model, we have that the values of abstained labels for which the previous optimisation problems are solved are, respectively
\begin{equation}
	\setlength{\abovedisplayskip}{6pt}
	\setlength{\abovedisplayshortskip}{6pt}
	\setlength{\belowdisplayskip}{6pt}
	\setlength{\belowdisplayshortskip}{6pt}	
	\hat{\underline{\vect{y}}}_{
	 \setindices_{\mathcal{A}}^{j\!-\!1}} \!\!:=\!
	 \argmax_{\!\!\scaleto{\vect{y}\in
	 	\{0,1\}^{\left|\setindices_{\mathcal{A}}^{j-1}\right|}}{7pt}}
		\!\hspace{-2.6mm}\prod_{~y_i\in\vect{y}}\!
		\!\!\frac{n(y_i|y_j\eq0)\!+\!s}{n(y_i|y_j\eq1)}
	\text{~and~~}
	\hat{\overline{\vect{y}}}_{
		\setindices_{\mathcal{A}}^{j\!-\!1}} \!\!:=\!
	  \argmin_{\!\scaleto{\vect{y}\in
	   	\{0,1\}^{\left|\setindices_{\mathcal{A}}^{j-1}\right|}}{7pt}} 
		\!\hspace{-2.6mm}\prod_{~y_i\in\vect{y}}\!
		\!\!\frac{n(y_i|y_j\eq0)}{n(y_i|y_j\eq1)\!+\!s}
	\!\!\label{eq:lowerupperpath}
\end{equation}
where $\setindices_{\mathcal{A}}^{j-1}$ is the set of indices of the $(j-1)th$ first predicted abstained labels, $n(\cdot)$ is a count function that counts the number of occurrences of the event $y_i|y_j$ and $n(y_i|y_j\eq1)$ is always strictly positive.
\end{proposition}

Proposition~\ref{prop:nccib} says that it is not necessary to know the original input features $\inputspace$ and neither the $(j-1)th$ first precise predicted labels, in order to get the lower and upper probability bound of Equations~\eqref{eq:impbranching}. However, it is necessary to keep track of the estimates made on all abstained labels, which is consistent with the fact that we want to capture the optimal lower and upper bounds of the conditional probability over all possible paths on which we have abstained.


Proposition~\ref{prop:nccib} allows us to propose an algorithm below that can calculate Equations \eqref{eq:lowerupperpath} linearly in the number of abstained labels.

\begin{proposition} \label{prop:ibcomplexity}
The chain of labels $\hat{\underline{\vect{y}}}_{\setindices_{\mathcal{A}}^{j-1}}$ and $\hat{\overline{\vect{y}}}_{\setindices_{\mathcal{A}}^{j-1}}$ can be obtained in a time complexity of $\mathcal{O}(|\setindices_{\mathcal{A}}^{j-1}|)$. 
\end{proposition}

The following proposition provides the time complexity of the inference step of the imprecise branching strategy, jointly with the NCC and previous results.

\begin{proposition}\label{prop:globalcomplexity}
	The global time complexity of the {\sc imprecise branching} strategy in the worst-case is $\mathcal{O}(m^2)$ and in the best-case  is $\mathcal{O}(m)$.
\end{proposition}

\subsection{Marginalization}
When the NCC is considered, nothing needs to be optimized in the marginalization strategy, thanks to the assumption of independence applied between the binary conditional models of the chain. 

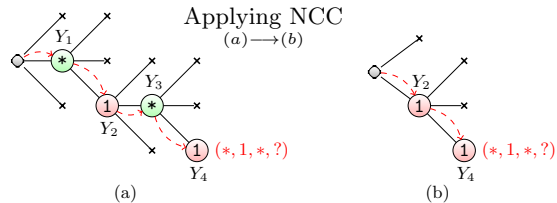
\begin{figure}[!th]
    \vspace{-2mm}\centering
	$\underset{(a) \longrightarrow (b)}{\text{\footnotesize Applying NCC}}$\\[-5ex]
	\resizebox{0.7\linewidth}{!}{
	  	\addtolength{\subfigcapskip}{-2mm}
		\subfigure[]{\label{fig:margfiga}
			\input{images/marginalization_general_from}
			\vspace{-6mm}
		}%
		\subfigure[]{
			\input{images/marginalization_general_from_ncc}
			\vspace{-4mm}
		}
	}\vspace{-4mm}
	\caption{Marginalization strategy applied to NCC for four labels $\{Y_1, Y_2, Y_3, Y_4\}$}
	\label{fig:marginalizationchain}\vspace{-2mm}
\end{figure}

We recall that the marginalization strategy needs to compute the conditional models described in Equations~\eqref{eq:marginalization}. These latter can be solved by simply ignoring the abstained labels in 
 Equations~\eqref{eq:ncclower} and \eqref{eq:nccupper} of the NCC. 
We thus focus on adapting Equation~\eqref{eq:nccupper} (Equation~\eqref{eq:ncclower} can be treated similarly), in order to show that the abstained labels can be removed of the conditioning and to get the expression presented in Equation~\eqref{eq:marupper2}.
Based on Equation~\eqref{eq:nccupper}, we can only focus on the conditional upper probability on labels, namely Equation~\eqref{eq:ncclabelupper}, and rewrite it as follows:
\begin{equation}\label{eq:lowercondmarg}
\setlength{\abovedisplayskip}{6pt}
\setlength{\abovedisplayshortskip}{6pt}
\setlength{\belowdisplayskip}{6pt}
\setlength{\belowdisplayshortskip}{6pt}
\overline{P}_{0}(\mathbf{Y}_{\setindices^{j-1}}=
		\hat{\vect{y}}_{\setindices^{j-1}})
		:= \overline{P}_{0}(\mathbf{Y}_{\setindices^{j-1}_*}=
		\hat{\vect{y}}_{\setindices^{j-1}_*},
		\mathbf{Y}_{\setindices^{j-1}_{\mathcal{A}}} = \{0, 1\}^{|\setindices_{\mathcal{A}}^{j-1}|})
\end{equation}
where $\setindices^{j-1}_*=\setindices_{\mathcal{R}}^{j-1}\cup\setindices_{\mathcal{I}}^{j-1}$ is the set of indices of relevant and irrelevant inferred labels, and the right side of last equation can be stated as 
\begin{equation*}
\setlength{\abovedisplayskip}{6pt}
\setlength{\abovedisplayshortskip}{6pt}
\setlength{\belowdisplayskip}{6pt}
\setlength{\belowdisplayshortskip}{6pt}
	\max_{\substack{P\in\left\{
		\credal_{Y_k|Y_j},
		\credal_{Y_a|Y_j}\right\}\\
		k\in\setindices^{j-1}_*,
		a\in\setindices^{j-1}_{\mathcal{A}}}} 
	\prod_{k\in\setindices^{j-1}_*} P(Y_k=\hat{y}_k|Y_{j}\eq0)
	\prod_{a\in\setindices^{j-1}_{\mathcal{A}}} P(Y_a=0 \cup Y_a = 1|Y_{j}\eq0).
\end{equation*}
Thanks to the product assumption in the NCC, each term can be treated separately, making it possible to decouple the multiplication in two parts;
\begin{equation*}
	 \setlength{\abovedisplayskip}{6pt}
	 \setlength{\abovedisplayshortskip}{6pt}
	 \setlength{\belowdisplayskip}{6pt}
	 \setlength{\belowdisplayshortskip}{6pt}
	 \max_{\substack{P\in\credal_{Y_k|Y_j}\\k\in\setindices^{j-1}_*}} 
		 \prod_{k\in\setindices^{j-1}_*} P(Y_k=\hat{y}_k|Y_{j}\eq0)
	 \times
	 \max_{\substack{P\in\credal_{Y_a|Y_j}\\
			a\in\setindices^{j-1}_{\mathcal{A}}}}
		\prod_{a\in\setindices^{j-1}_{\mathcal{A}}} P(Y_a=0 \cup Y_a = 1|Y_{j}\eq0),
\end{equation*}
where $P(Y_a\!\eq 0\cup Y_a\!\eq 1|Y_{j}\!\eq 0)\!\eq\!1$, and hence the second part becomes $1$. Replacing this result in Equation~\eqref{eq:lowercondmarg}, and then this latter in Equation~\eqref{eq:nccupper}, we get
\begin{equation*}
  \setlength{\abovedisplayskip}{6pt}
  \setlength{\abovedisplayshortskip}{6pt}
  \setlength{\belowdisplayskip}{6pt}
  \setlength{\belowdisplayshortskip}{6pt}	
  \uppercondprobchain(Y_j=1) =
	\max\nolimits_{P\in\credal_{Y_j|
	Y_{\subidx{R}}, 
	Y_{\subidx{I}}}} 
	P_{\newinstance}(Y_j=1|
			Y_{\setindices_\mathcal{R}^{j-1}}=1,
			Y_{\setindices_{\mathcal{I}}^{j-1}}=0).
\end{equation*}
Therefore, at each inference step, we can apply Equations \eqref{eq:ncclower} and \eqref{eq:nccupper} on the reduced new formulation of the marginalization strategy \hyperref[eq:marginalization2]{(MAR*)}. An illustration providing some intuition about this reduction can be seen in Figure~\ref{fig:marginalizationchain}.
\section{Experiments}\label{sec:experimentschain}
In this section, we perform experiments\footnote{Implemented in Python, see \url{https://github.com/sdestercke/classifip}} on 6 data sets issued from the MULAN repository\footnote{\url{http://mulan.sourceforge.net/datasets.html}} (c.f. Table~\ref{tab:datasetschain}), following a $10\!\times\!10$ cross-validation procedure.
\begin{table}[!ht]
	\centering\setlength{\tabcolsep}{5pt}
	\resizebox{0.85\columnwidth}{!}{%
	\begin{tabular}{ccccccc}
		 Data set & \#Domain & \#Features & \#Labels & \#Instances &
		 	\#Cardinality & \#Density \\
		\hline
		emotions & music & 72 & 6 & 593 & 1.90 & 0.31 \\
	 	scene & image  & 294 & 6 & 2407 & 1.07 & 0.18 \\
		yeast & biology & 103 & 14 & 2417 & 4.23 & 0.30 \\
		cal500 & music & 68 & 174 & 502 & 26.04 & 0.15\\
		medical & text & 1449 &  45 & 978 & 1.25 & 0.03	\\
		enron & text & 1001 & 53 & 1702 & 3.38 & 0.06	
	\end{tabular}}\vspace{2mm}
	\caption{Multi-label data sets summary}\label{tab:datasetschain}
	\vspace{-8mm}
\end{table}
\subsubsection*{Evaluation and Setting}
The usual metrics used in multi-label problems are not adapted at all when we infer set-valued predictions. Thus, we consider appropriate to use the set-accuracy (SA) and completeness (CP) \cite[\S 4.1]{destercke2014multilabel}, as follows
\begin{equation*}
\setlength{\abovedisplayskip}{4pt}
\setlength{\abovedisplayshortskip}{4pt}
\setlength{\belowdisplayskip}{4pt}
\setlength{\belowdisplayshortskip}{4pt}
	SA(\hat{\bm y}, \bm y) = \mathbbm{1}_{(\bm y \in \hat{\bm y})} 
 \quad\text{and}\quad
	CP(\hat{\bm y}, \bm y) = \frac{|Q|}{m},
\end{equation*}
where $\hat{\bm y}$ is the partial binary prediction (i.e. the set of all possibles binary vectors) and $Q$ denote the set of non-abstained labels. When predicting complete vectors, then $CP=\!1$ and $SA$ equals the 0/1 loss function and when predicting the empty vector, i.e. all labels $\hat{y}_i=*$, then $CP=\!0$ and by convention $SA=\!1$. The reason for using $SA$ is that chaining is used as an approximation of the optimal prediction for a 0/1 loss function.
\paragraph*{Imprecise Classifier}
As mentioned in Section~\ref{sec:mlcncc}, we will use the \emph{naive credal classifier} (NCC). Note that NCC needs discretized input spaces, so we discretize data sets to $z\eq6$ intervals (except for Medical and Enron data sets). Besides, we restrict the values of the hyper-parameter of the imprecision to $s\in\{0.0, 0.5, \dots, 4.5, 5.5\}$ (when $s\eq0.0$, NCC becomes the precise classifier NBC). At higher values of $s$, the NCC model will make mostly vacuous predictions (i.e. abstain in all labels $\forall i, Y_i\eq*$) for the data sets we consider here.

\paragraph*{Missing Labels}
To simulate missingness during the training step, we uniformly pick at random a percentage of labels $Y_{j,i}$ (the $j$th label of the $i$th instance), which are then removed from the training data used to fit the conditional models in the chain. In this paper, we set up five different percentages of missingness: $\{0, 20, 40, 60, 80\}$\%.

\subsection{Experimental Results}
\begin{figure}[!th]
	\vspace{-4mm}\centering
	\subfigure[\sc Emotions]{\hspace{-2mm}
	   \includegraphics[width=0.33\linewidth]{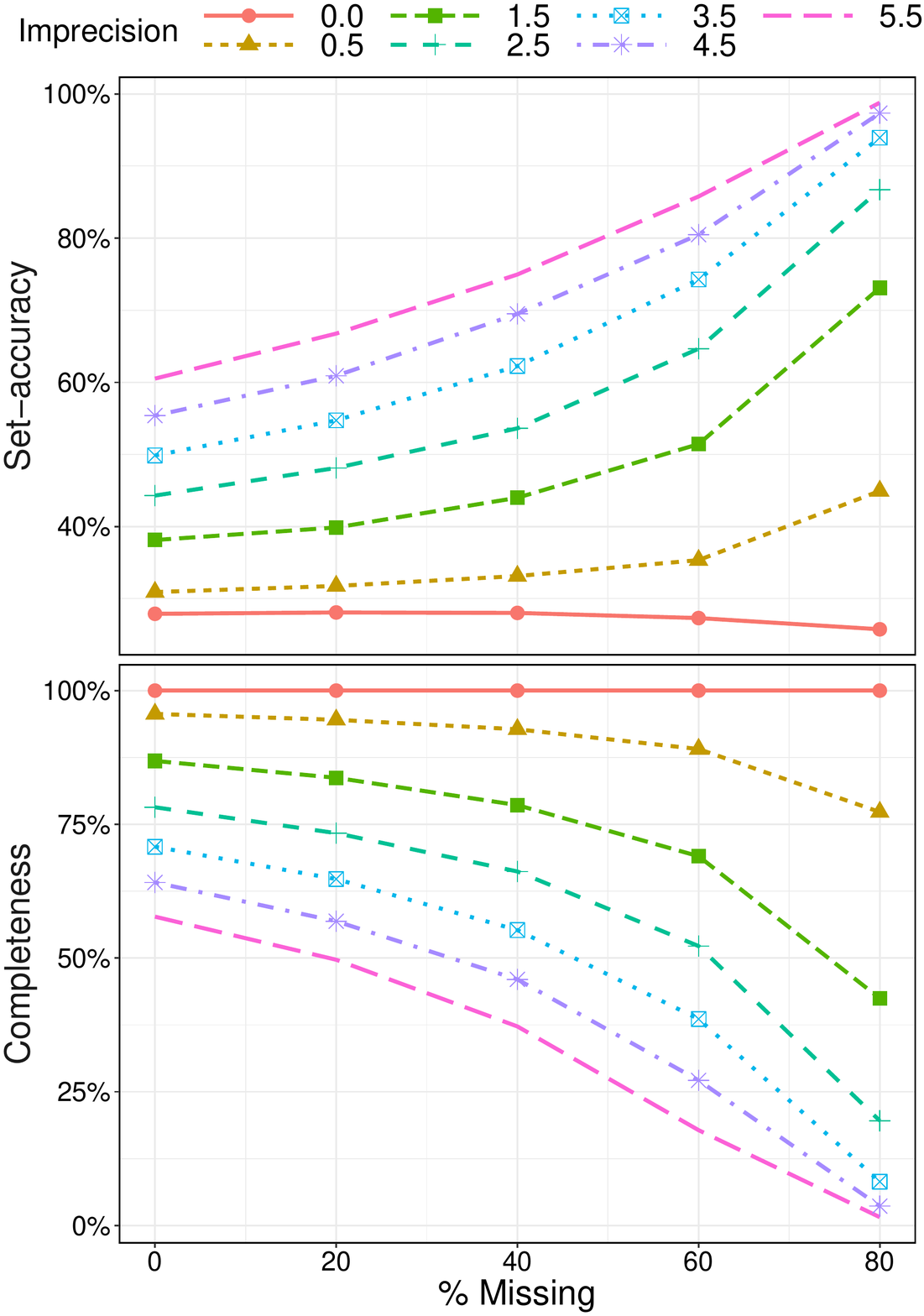}
	}%
	\subfigure[\sc Enron]{\hspace{-1mm}
		\includegraphics[width=0.33\linewidth]{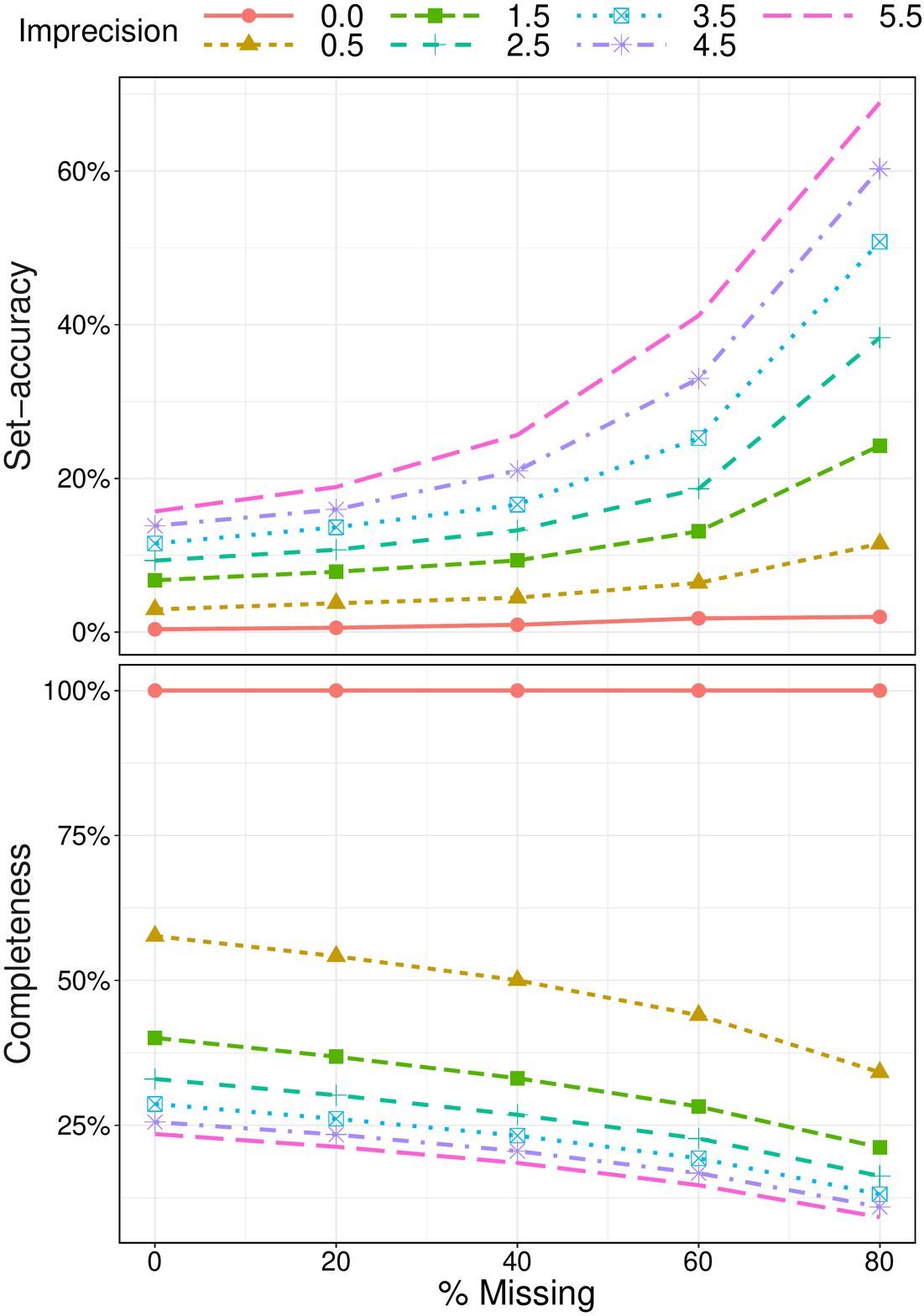}
	}%
	\subfigure[\sc Yeast]{\hspace{-1mm}
		\includegraphics[width=0.33\linewidth]{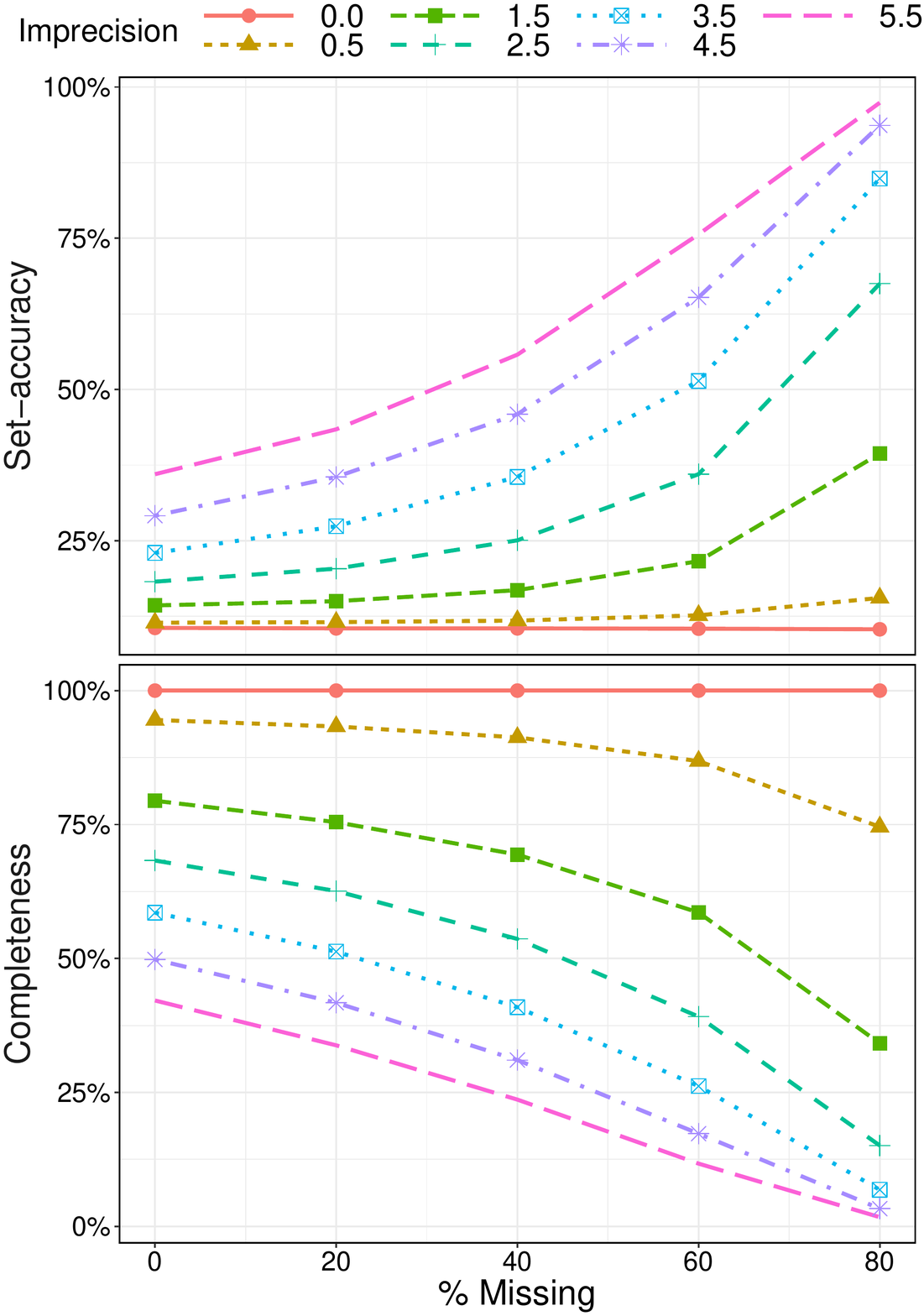}
	}%
	\vspace{-2mm}\caption{\textbf{Missing labels - Imprecise Branching} Evolution of the average (\%) set-accuracy (top) and completeness (down) for each level of imprecision (a curve for each one) and a discretization $z=6$, with respect to the percentage of missing labels.}\label{fig:expmissibaset}\vspace{-4mm}%
\end{figure}

Figure~\ref{fig:expmissibaset}, 
we provide the results for 3 data sets, showing set-accuracy and completeness measures in average\footnote{The confidence intervals obtained on the experimental results are very small and we therefore prefer not to display them in the figures in order not to overcharge them.}(\%) obtained by fitting the NCC model for different percentages of missing labels, respectively, applied to the data sets of Table~\ref{tab:datasetschain} and using the imprecise branching strategy (trends were similar for the marginalization strategies, not displayed due to lack of space)\footnote{The supplementary results can be found in the online extend version~\cite{alarcon2021multilabel}}.

The results show that when the percentage of missing labels increases, the set-accuracy ($SA$) increases (regardless of the amount of imprecision $s$ we inject) as we abstain more and more (as completeness decreases). This means that the more imprecise we get, the more accurate are those predictions we retain. However, a high amount of imprecision is sometimes required to include the ground-truth solution within the set-valued prediction (this may be due to the very restrictive 0/1 loss metric). For instance, with $s\!\eq\!5.5$ and $40\%$ of missingness, we get a $\!>\!65\%$ of set-accuracy versus a $\!<\!50\%$ of completeness, in the Emotions data set.

Overall, the results are those that we expect and are also sufficient to show that cautious inferences with probability sets may provide additional benefits when dealing with missing labels.

\section{Conclusions}
In this paper, we have proposed two new strategies to adapt the classical chaining multi-label problem to the case of handling imprecise probability estimates. Such strategies come with daunting challenges to obtain cautious and reliable predictions, and have been successfully resolved using the NCC model. 

While the NCC makes easy to solve the strategies thanks to its assumptions, the same restrictive assumptions may also be the reason why the initial accuracy is rather low (especially for Yeast). Indeed, it seems reasonable to think that the independence assumptions somehow limit the benefits of including label dependencies information through the chaining. It seems therefore essential, in future works, to investigate other classifiers as well as to solve optimisation issues in a general or approximative way.

Another open issue is how we can use or extend the existing heuristics of probabilistic classifier approaches on our proposal strategies, such as epsilon-approximate inference, $A^*$ and beam search methods~\cite{mena2016overview,kumar2013beam}. 

\subsubsection*{Acknowledgments.} We wish to thank the reviewers of the ECSQARU conference, especially one of them who was thorough, and made us rethink a lot of aspects (of which only some are visible in the revised version).

%

\bibliographystyle{splncs04}
\bibliography{bibliography}

\input{suplementary_proofs}

\end{document}

%% file: images/precise_chaining.tex
\tikzset{
  treenode/.style = {shape=rectangle, rounded corners,
                     draw, align=center,
                     top color=white, bottom color=blue!20},
  root/.style     = {treenode, font=\Large, bottom color=red!30},
  env/.style      = {treenode, font=\ttfamily\normalsize},
  dummy/.style    = {circle,draw}
}
\tikzstyle{level 1}=[level distance=3.8cm, sibling distance=2.2cm]
\tikzstyle{level 2}=[level distance=4.5cm, sibling distance=1cm]
\tikzstyle{bag} = [text width=4em, text centered]
\tikzstyle{end} = [circle, minimum width=3pt,fill, inner sep=0pt]
\begin{tikzpicture}[
	grow=right, 
	sloped,
	edge from parent/.style = {draw, -latex,
		 font=\footnotesize\sffamily, text=blue},
	every node/.style  = {font=\footnotesize}
	]
\node[root] {}
    child {
        node[env] {$Y_1=1$}        
            child {
                node[end, label={[red]right:{
              	{$\bf (1, 1)$}}}] {}
                edge from parent
                node[below]{
                $P_{\newinstance}(Y_2\!=\!1|Y_1\!=\!1)\!=\!
                \textcolor{red}{\bf 0.6}$}
            }
            child {
                node[end, label=right:{$(1, 0)$}] {}
                edge from parent
                node[above]{$P_{\newinstance}(Y_2\!=\!0|Y_1\!=\!1)\!=\! 
                \textcolor{red}{\bf 0.4}$}
            }
            edge from parent 
            node[below] {$P_{\newinstance}(Y_1=1)\!=\!
            \textcolor{red}{\bf 0.6}$}
    }
    child {
        node[env] {$Y_1=0$}       
        child {
              node[end,label=right:{$(0, 1)$}] {}
              edge from parent
              node[below]{$P_{\newinstance}(Y_2\!=\!1|Y_1\!=\!0)\!=\! 
               \textcolor{red}{\bf 0.1}$}
            }
            child {
                node[end, label=right:{$(0, 0)$}] {}
                edge from parent
                node[above]{
                $P_{\newinstance}(Y_2\!=\!0|Y_1\!=\!0)\!=\!
                \textcolor{red}{\bf 0.9}$}
            }
        edge from parent         
            node[above] {$P_{\newinstance}(Y_1=0)=
            \textcolor{red}{\bf 0.4}$}
    };
\end{tikzpicture}

%% file: images/precise_chaining_inverse.tex
\tikzset{
  treenode/.style = {shape=rectangle, rounded corners,
                     draw, align=center,
                     top color=white, bottom color=blue!20},
  root/.style     = {treenode, font=\Large, bottom color=red!30},
  env/.style      = {treenode, font=\ttfamily\normalsize},
  dummy/.style    = {circle,draw}
}
\tikzstyle{level 1}=[level distance=3.8cm, sibling distance=2.2cm]
\tikzstyle{level 2}=[level distance=4.5cm, sibling distance=1cm]
\tikzstyle{bag} = [text width=4em, text centered]
\tikzstyle{end} = [circle, minimum width=3pt,fill, inner sep=0pt]
\begin{tikzpicture}[
	grow=right, 
	sloped,
	edge from parent/.style = {draw, -latex, font=\tiny\sffamily, text=blue},
	every node/.style  = {font=\footnotesize}
	]
\node[root] {}
    child {
        node[env] {$Y_2=1$}        
            child {
                node[end, label=right:{$(1, 1)$}] {}
                edge from parent
                node[below]{
                $P_{\newinstance}(Y_1\!=\!1|Y_2\!=\!1)\!=\!
                \textcolor{red}{\bf 0.9}$}
            }
            child {
                node[end, label=right:{$(0,1)$}] {}
                edge from parent
                node[above]{
                $P_{\newinstance}(Y_1\!=\!0|Y_2\!=\!1)\!=\! 
                \textcolor{red}{\bf 0.1}$}
            }
            edge from parent 
            node[below] {$P_{\newinstance}(Y_2=1)\!=\!
            \textcolor{red}{\bf 0.4}$}
    }
    child {
        node[env] {$Y_2=0$}       
        child {
              node[end,label=right:{$(1,0)$}] {}
              edge from parent
              node[below]{
               $P_{\newinstance}(Y_1\!=\!1|Y_2\!=\!0)\!=\! 
               \textcolor{red}{\bf 0.4}$}
            }
            child {
                node[end, label={[red]right:{
              	{$\bf (0, 0)$}}}] {}
                edge from parent
                node[above]{
                $P_{\newinstance}(Y_1\!=\!0|Y_2\!=\!0)\!=\!
                \textcolor{red}{\bf 0.6}$}
            }
        edge from parent         
            node[above] {$P_{\newinstance}(Y_2=0)=
            \textcolor{red}{\bf 0.6}$}
    };
\end{tikzpicture}

%% file: images/mix_precise_imprecise_chaining.tex
\tikzset{
  treenode/.style = {shape=rectangle, rounded corners,
                     draw, align=center,
                     top color=white, bottom color=blue!20},
  root/.style     = {treenode, font=\Large, bottom color=red!30},
  env/.style      = {treenode, font=\ttfamily\normalsize},
  dummy/.style    = {circle,draw}
}
\tikzstyle{level 1}=[level distance=4cm, sibling distance=2.0cm]
\tikzstyle{level 2}=[level distance=4cm, sibling distance=0.8cm]
\tikzstyle{bag} = [text width=4em, text centered]
\tikzstyle{end} = [circle, minimum width=3pt,fill, inner sep=0pt]
\begin{tikzpicture}[
	grow=right, 
	sloped,
	edge from parent/.style = {draw, -latex,
		 font=\footnotesize\sffamily, text=blue},
	every node/.style  = {font=\footnotesize}]
\node[root] {}
    child {
        node[env] {$Y_1=1$}        
            child[very thick] {
                node[end, label=right:{$(1, 1)$}] {}
                edge from parent
                node[below] {$\bf P_{\newinstance}(Y_2\!=\!1|Y_1\!=\!1)\!=\!0.6$}
            }
            child {
                node[end, label=right:{$(1, 0)$}] {}
                edge from parent
                node[above] {$P_{\newinstance}(Y_2\!=\!0|Y_1\!=\!1)\!=\!0.4$}
            }
            edge from parent 
            node[below] {
            $[\hat{P}_{\newinstance}](Y_1\!\!=\!\!1)[0.45, 0.7]$}
    }
    child {
        node[env] {$Y_1=0$}       
        child[very thick] {
                node[end,  label={right:{{$ (0, 1)$}}}] {}
                edge from parent
                node[below] {$\bf P_{\newinstance}(Y_2\!=\!1|Y_1\!=\!0)\!=\!0.1$}
            }
            child {
                node[end, label=right:{$(0, 0)$}] {}
                edge from parent
                node[above] {$P_{\newinstance}(Y_2\!=\!0|Y_1\!=\!0)\!=\!0.9$}
            }
        edge from parent         
            node[above] {
            $[\hat{P}_{\newinstance}](Y_1\!\!=\!\!0)[0.3, 0.55]$}
    };
\end{tikzpicture}

%% file: images/branching_chaining.tex
\tikzset{
  treenode/.style = {shape=rectangle, rounded corners,
                     draw, align=center,
                     top color=white, 
                     bottom color=blue!20},
  root/.style     = {treenode, 
  		font=\Large, bottom color=black!30},
  one/.style      = {treenode, 
  		  minimum size=4mm,
		 inner sep=0pt, outer sep=0pt,
  		 shape=circle,
  		 bottom color=red!30,
  		 top color=white,
  		 align=center,
  		 font=\ttfamily\normalsize},
  zero/.style      = {treenode, 
  		 minimum size=4mm,
		 inner sep=0pt, outer sep=0pt,
		 shape=circle,
		 top color=white,
  		 align=center,
  		 font=\ttfamily\normalsize},		 
  cross/.style={cross out, draw=black, solid, 
  	thick, minimum size=1.5*(#1-\pgflinewidth), 
  	inner sep=1.5, outer sep=0pt},
}
\tikzstyle{level 1}=[level distance=0.8cm, sibling distance=2.2cm]
\tikzstyle{level 2}=[level distance=0.8cm, sibling distance=2cm]
\tikzstyle{level 3}=[level distance=0.8cm, sibling distance=1cm]
\tikzstyle{level 4}=[level distance=0.8cm, sibling distance=1cm]
\tikzstyle{level 5}=[level distance=0.8cm, sibling distance=0.5cm]
\tikzstyle{bag} = [text width=4em, text centered]
\tikzstyle{end} = [circle, minimum width=3pt,fill, inner sep=0pt]
\tikzstyle{pruned} = [cross, minimum width=3pt,fill, inner sep=0pt]
\begin{tikzpicture}[
	grow=right, 
	sloped,
	edge from parent/.style = 
		{draw, font=\tiny\sffamily, text=blue},
	every node/.style = {font=\footnotesize}	]
\node(r)[root] {}
    child {
        node[cross] {}        
        edge from parent 
        node[below] {}
    }
    child {
        node(0)[zero] {0}       
        child {
            node(01)[one] {1}
            child {
            	node(011)[one] {1}
            	child {
	              node(0111)[one] {1}     
	            	child {
	              	  node(01111)[one, 
	              	  	label=right:{$(0, 1, 1, 1, 1)$}] {1}        	
	            	}
		            child {
		              node[end] {}        	
	                }                 	
	            }
	            child {
	              node(0110)[zero] {0}  
	              child {
	              	node(01101)[one,
	              		label=right:{$(0, 1, 1, 0, 1)$}] {1}         	
	           	  }
		          child {
		            node[end] {}        	
	              }       	
	            }
            }
            child {
            	node[cross] {}
            }
            edge from parent
            node[below] {}
        }
        child {
            node(00)[zero]{0}
            child {
            	node(001)[one] {1}
            	child {
	              node(0011)[one] {1}    
	              child {
	              	node(00111)[one,
	              		label=right:{$(0, 0, 1, 1, 1)$}] {1}         	
	           	  }
		          child {
		            node[end] {}        	
	              }      	
	            }
	            child {
	              node(0010)[zero] {0}        	
	              child {
	              	node(00101)[one,
	              		label=right:{$(0, 0, 1, 0, 1)$}] {1}
	           	  }
		          child {
		            node[end] {}        	
	              }  
	            }
            }
            child {
                node[cross] {}
            }
            edge from parent
            node[above] {}
         }
         edge from parent         
         node[above] {}
    };
\draw[dashed,->, bend left, red]  (r) to (0);
\draw[dashed,->, bend right, red] (0) to (01);
\draw[dashed,->, bend left, red]  (0) to (00);
\draw[dashed,->, bend right, red] (00) to (001);
\draw[dashed,->, bend right, red] (01) to (011);
\draw[dashed,->, bend right, red] (001) to (0011);
\draw[dashed,->, bend left, red]  (001) to (0010);
\draw[dashed,->, bend right, red] (011) to (0111);
\draw[dashed,->, bend left, red]  (011) to (0110);
\draw[dashed,->, bend right, red] (0010) to (00101);
\draw[dashed,->, bend right, red] (0011) to (00111);
\draw[dashed,->, bend right, red] (0111) to (01111);
\draw[dashed,->, bend right, red] (0110) to (01101);
\end{tikzpicture}

%% file: images/marginalization_general_from.tex
\tikzset{
  treenode/.style = {shape=rectangle, rounded corners,
                     draw, align=center,
                     top color=white, 
                     bottom color=blue!20},
  root/.style     = {treenode, 
  		font=\Large, bottom color=black!30},
  one/.style      = {treenode, 
  		  minimum size=4mm,
		 inner sep=0pt, outer sep=0pt,
  		 shape=circle,
  		 bottom color=red!30,
  		 top color=white,
  		 align=center,
  		 font=\ttfamily\normalsize},
  zero/.style      = {treenode, 
  		 minimum size=4mm,
		 inner sep=0pt, outer sep=0pt,
		 shape=circle,
		 top color=white,
  		 align=center,
  		 font=\ttfamily\normalsize},	
  imprecise/.style      = {treenode, 
  		 minimum size=4mm,
		 inner sep=0pt, outer sep=0pt,
		 shape=circle,
		 bottom color=green!30,
		 top color=white,
  		 align=center,
  		 font=\ttfamily\normalsize},		 
  cross/.style={cross out, draw=black, solid, 
  	thick, minimum size=1.5*(#1-\pgflinewidth), 
  	inner sep=1.5, outer sep=0pt},
}
\tikzstyle{level 1}=[level distance=0.8cm, sibling distance=0.8cm]
\tikzstyle{level 2}=[level distance=0.8cm, sibling distance=0.8cm]
\tikzstyle{level 3}=[level distance=0.8cm, sibling distance=0.8cm]
\tikzstyle{level 4}=[level distance=0.8cm, sibling distance=0.8cm]
\tikzstyle{level 5}=[level distance=0.8cm, sibling distance=0.8cm]
\tikzstyle{bag} = [text width=4em, text centered]
\tikzstyle{end} = [circle, minimum width=3pt,fill, inner sep=0pt]
\tikzstyle{pruned} = [cross, minimum width=3pt,fill, inner sep=0pt]
\begin{tikzpicture}[
	grow=right, 
	sloped,
	edge from parent/.style = 
		{draw, font=\tiny\sffamily, text=blue},
	every node/.style = {font=\footnotesize}	]
\node(r)[root] {}
    child {
        node (1)[cross] {}        
        edge from parent 
        node[below] {}
    }
    child {
        node(2)[imprecise]{*}
        child {
            node(21)[one] {1}
            child {
            	node(211)[cross] {}
            }
            child {
        		node(212)[imprecise]{*}
        		child {
	              node(2121)[one,
	              		label=right:{\tred{\bf $(*, 1, *, ?)$}}] {1}    
	            }
	            child {
     			   node(2122)[cross]{}
			    }
	            child {
	              node(2120)[cross] {}
	            }
    		}
            child {
           		node(210)[cross]{}
            	edge from parent
            	node[above] {}
           }
        }
        child {
            node(02)[cross]{}
        }
        child {
            node(00)[cross]{}
            edge from parent
            node[above] {}
        }
        edge from parent         
        node[above] {}
    }
    child {
        node(0)[cross]{}
    };
\draw[dashed,->, bend left, red]  (r) to (2);
\draw[dashed,->, bend left, red] (2) to (21);
\draw[dashed,->, bend right, red] (21) to (212);
\draw[dashed,->, bend right, red] (212) to (2121);

\node[yshift=12pt, xshift=1pt] at (2) {$Y_1$};
\node[yshift=-12pt, xshift=1pt] at (21) {$Y_2$};
\node[yshift=12pt, xshift=1pt] at (212) {$Y_3$};
\node[yshift=-12pt, xshift=1pt] at (2121) {$Y_4$};
\end{tikzpicture}

%% file: images/marginalization_general_from_ncc.tex
\tikzset{
  treenode/.style = {shape=rectangle, rounded corners,
                     draw, align=center,
                     top color=white, 
                     bottom color=blue!20},
  root/.style     = {treenode, 
  		font=\Large, bottom color=black!30},
  one/.style      = {treenode, 
  		  minimum size=4mm,
		 inner sep=0pt, outer sep=0pt,
  		 shape=circle,
  		 bottom color=red!30,
  		 top color=white,
  		 align=center,
  		 font=\ttfamily\normalsize},
  zero/.style      = {treenode, 
  		 minimum size=4mm,
		 inner sep=0pt, outer sep=0pt,
		 shape=circle,
		 top color=white,
  		 align=center,
  		 font=\ttfamily\normalsize},	
  imprecise/.style      = {treenode, 
  		 minimum size=4mm,
		 inner sep=0pt, outer sep=0pt,
		 shape=circle,
		 bottom color=green!30,
		 top color=white,
  		 align=center,
  		 font=\ttfamily\normalsize},		 
  cross/.style={cross out, draw=black, solid, 
  	thick, minimum size=1.5*(#1-\pgflinewidth), 
  	inner sep=1.5, outer sep=0pt},
}
\tikzstyle{level 1}=[level distance=0.8cm, sibling distance=1.2cm]
\tikzstyle{level 2}=[level distance=0.8cm, sibling distance=0.8cm]
\tikzstyle{level 3}=[level distance=0.8cm, sibling distance=0.8cm]
\tikzstyle{level 4}=[level distance=0.8cm, sibling distance=0.8cm]
\tikzstyle{level 5}=[level distance=0.8cm, sibling distance=0.8cm]
\tikzstyle{bag} = [text width=4em, text centered]
\tikzstyle{end} = [circle, minimum width=3pt,fill, inner sep=0pt]
\tikzstyle{pruned} = [cross, minimum width=3pt,fill, inner sep=0pt]
\begin{tikzpicture}[
	grow=right, 
	sloped,
	edge from parent/.style = 
		{draw, font=\tiny\sffamily, text=blue},
	every node/.style = {font=\footnotesize}	]
\node(r)[root] {}
    child {
        node(1)[one] {1}        
        child {
           node(11)[one, label=right:{\tred{\bf $(*, 1, *, ?)$}}] {1}
        }
        child {
        	node(12)[cross]{}
    	}
        child {
           node(10)[cross]{}
           edge from parent
           node[above] {}
        }
    }
    child {
        node(0)[cross] {}  
    };
\draw[dashed,->, bend left, red]  (r) to (1);
\draw[dashed,->, bend left, red] (1) to (11);
\node[yshift=12pt, xshift=1pt] at (1) {$Y_2$};
\node[yshift=-12pt, xshift=1pt] at (11) {$Y_4$};

\end{tikzpicture}

%% file: suplementary_proofs.tex
\newpage
\appendix
\section*{Appendix}
\section{NCC derivation for multi-label chaining}
As the purpose of the imprecise chaining is to compute binary conditional dependence models, we need only get conditional probability bounds of the probability $P(Y_{j}=y_j|\mathbf{X}=\newinstance, Y_{\setindices^{j-1}}\!\!=\hat{\vect{y}}_{\setindices^{j-1}})$, so by using Bayes' theorem and naive Bayes' attribute independence assumption, it can be written as follows
\begin{align}\label{eq:marginalbayeschain}\!\!
\frac{P(Y_{j}=y_j)
	\prod_{i=1}^p  P(X_i = x_i|Y_{j}=y_j)
	\prod_{k=1}^{j-1} 
		P(Y_{k}\!\!=\hat{y}_{k}|Y_{j}=y_j)
}{\sum_{y_l\in\{0, 1\}}P(Y_{j}=y_l)
	\prod_{i=1}^p  P(X_i = x_i|Y_{j}=y_l)
	\prod_{k=1}^{j-1} 
		P(Y_{k}\!\!=\hat{y}_{k}|Y_{j}=y_l)
}.
\end{align}

Computing lower and upper probability bounds $[\underline{P},\overline{P}]$ over all possible marginals $\credal_{Y_j}$ and conditional distributions $\credal_{X_i|Y_j}, \credal_{Y_k|Y_j}$ can be performed by solving the following minimization/maximization problem of Equation~\eqref{eq:marginalbayeschain} as follows
\begin{align}
\underline{P}(Y_{j}=y_j|&X\!\!=\!\newinstance, Y_{\setindices^{j-1}}\!\!=\!\hat{\bm{y}}_{\setindices^{j-1}}) =\nonumber\\
&\min_{P\in\credal_{Y_j}} \min_{\substack{P\in\left\{\credal_{X_i|Y_j},\credal_{Y_k|Y_j}\right\}\\i=1,\dots,d\\k=1,\dots, j-1}} P(Y_{j}=y_j|X\!\!=\!\newinstance,Y_{\setindices^{j-1}}\!\!=\!\hat{\bm{y}}_{\setindices^{j-1}}),\\
\overline{P}(Y_{j}=y_j|&X\!\!=\!\newinstance,Y_{\setindices^{j-1}}\!\!=\!\hat{\bm{y}}_{\setindices^{j-1}}) =\nonumber\\
&\max_{P\in\credal_{Y_j}} \max_{\substack{P\in\left\{\credal_{X_i|Y_j},\credal_{Y_k|Y_j}\right\}\\i=1,\dots,d\\k=1,\dots, j-1}} P(Y_{j}=y_j|X\!\!=\!\newinstance,Y_{\setindices^{j-1}}\!\!=\!\hat{\bm{y}}_{\setindices^{j-1}}).
\end{align}

In practice, we assume a precise estimation of the marginal distribution $\mathbb{P}_{Y_j}$ in lieu of a credal set $\credal_{Y_j}$, so  optimization problems over the credal set of marginal distributions $\credal_{Y_j}$ can be ignored. Therefore, one can easily show that last equations evaluated to $Y_{j}=1$ ($Y_{j}=0$ can be directly calculated using duality) are equivalent to 
\begin{align}
	\underline{P}(Y_{j}\eq1|
		\mathbf{X}\eq\newinstance, 
		Y_{\setindices^{j-1}}=\hat{\bm{y}}_{\setindices^{j-1}}) &\eq
		\left(\!1+ \frac{P(Y_{j}\eq0)
				\overline{P}_{0}(\mathbf{X}\eq{\newinstance})
				\overline{P}_{0}(Y_{\setindices^{j-1}}\eq
					\hat{\bm{y}}_{\setindices^{j-1}})
			}
			{P(Y_{j}\eq1)
				\underline{P}_{1}(\mathbf{X}\eq{\newinstance})
				\underline{P}_{1}(Y_{\setindices^{j-1}}\eq
					\hat{\bm{y}}_{\setindices^{j-1}})
			}
		\!\right)^{-1}\!\!\! \label{eq:ncclowerproof} \\
	\!\!\!\overline{P}(Y_{j}\eq1| \mathbf{X}
		\eq\newinstance, Y_{\setindices^{j-1}}\eq
			\hat{\bm{y}}_{\setindices^{j-1}}) &\eq
		\left(\!1+ \frac{P(Y_{j}=0)
			\underline{P}_{0}(\mathbf{X}\eq{\newinstance})
				\underline{P}_{0}(Y_{\setindices^{j-1}}\eq
					\hat{\bm{y}}_{\setindices^{j-1}})
			}
			{P(Y_{j}\eq1)
				\overline{P}_{1}(\mathbf{X}={\newinstance})
					\overline{P}_{1}(Y_{\setindices^{j-1}}\eq
						\hat{\bm{y}}_{\setindices^{j-1}})
			}
		\!\right)^{-1}\!\!\!\!\! \label{eq:nccupperproof}
\end{align}
where probability bounds $[\underline{P}_{1}, \overline{P}_{1}]$ and $[\underline{P}_{0}, \overline{P}_{0}]$ of each different conditional event are defined as follows
{\small\begin{align}
	\overline{P}_{0}(\mathbf{X}=\newinstance)\!:=\!
		\prod_{i=1}^p \overline{P}(X_i = x_i|Y_{j}\eq0)
	~~&\text{and}~~
	\overline{P}_{0}(\mathbf{Y}_{\setindices^{j-1}}=
			\vect{y}_{\setindices^{j-1}})\!:=\!
		\prod_{k=1}^{j-1} \overline{P}(Y_k=\hat{y}_k|Y_{j}\eq0),
	\label{eq:ncclabelupperproof}
	\\
	\underline{P}_{1}(\mathbf{X}=\newinstance)\!:=\!
		\prod_{i=1}^p \underline{P}(X_i = x_i|Y_{j}\eq1)
	~~&\text{and}~~
	\underline{P}_{1}(\mathbf{Y}_{\setindices^{j-1}}=
			\vect{y}_{\setindices^{j-1}})\!:=\!
		\prod_{k=1}^{j-1} \underline{P}(Y_k=\hat{y}_k|Y_{j}\eq1).
	\label{eq:ncclabellowerproof}
\end{align}}

The last conditional probability bounds are derived using the Imprecise Dirichlet model (IDM)~\cite{walley1996inferences}
\begin{align}\label{eq:nccidm}
	\underline{P}(X_i\eq x_i| Y_{j}\eq y_j)\eq  
		\frac{n(x_i|y_j)}{n(y_j)+s}
	~~\text{and}~~
	\overline{P}(X_i\eq x_i| Y_{j}\eq y_j)\eq 
		\frac{n(x_i|y_j)+s}{n(y_j)+s}
\end{align}
where $n(\cdot)$ is a count function that counts the number of occurrences of events $x_i|y_j$ and $y_j$ in the observed data set. For instance; $n(x_i|y_j)$ is the number of instances in the training set where $X_i=x_i$ and the label value is $Y_i=y_i$. 

In the same way as precedent equations, we can obtain probability bounds $[\underline{P}(Y_k=\hat{y}_k|Y_{j}=y_j), \overline{P}(Y_k=\hat{y}_k|Y_{j}=y_j)]$.

\begin{remark}
	The denominator of Equations \eqref{eq:ncclowerproof} and \eqref{eq:nccupperproof} may in some cases (depending on the training data) become zero, and therefore, the division would not be defined. In these cases, we adopt the ``Laplace smoothing'' technique for the precise probabilities $P(Y_{j}=0)$ and $P(Y_{j}=1)$. Furthermore, if we obtain an undefined division, i.e. $0/0$, we manually put $0$.
\end{remark}

\begin{remark}
	Note that the choice of the marginal distribution $\mathbb{P}_{Y_j}$ in lieu of a credal set $\credal_{Y_j}$ does not change the theoretical results obtained in Proposition~\ref{prop:nccib}. However, if the credal $\mathscr{P}_Y$ is considered, we can obtain a very small, but not significant, improvement in the experimental results\footnote{Anyone can reproduce these results by using our online implementation.}.
\end{remark}

\section{Proofs for our results}
\begin{proof}[\textbf{of Proposition~\ref{prop:nccib}}]
Let us begin to prove the optimization problem of the lower probability of \eqref{eq:impbranching} evaluated to $Y_j=1$
\begin{equation}
\lowercondprobchain(Y_j=1) = 
\min_{\vect{y}\in\{0,1\}^{\left|\setindices_{\mathcal{A}}^{j-1}\right|}} 
	\underline{P}_{\newinstance}(Y_j=1|
		Y_{\setindices_\mathcal{R}^{j-1}}=1, 
		Y_{\setindices_{\mathcal{I}}^{j-1}}=0,
		Y_{\setindices_{\mathcal{A}}^{j-1}}=\vect{y}).
\end{equation}
Let us to define $\setindices^{j-1}_*=\setindices_{\mathcal{R}}^{j-1}\cup\setindices_{\mathcal{I}}^{j-1}$ as the set of indices of relevant and irrelevant predicted labels down to the $(j-1){th}$ index. By applying Equation~\eqref{eq:ncclowerproof} to the right side of last equation, we get 
\begin{equation*}
	\min_{\vect{y}\in\{0,1\}^{\left|\setindices_{\mathcal{A}}^{j-1}\right|}} 
	\left(1+ \frac{P(Y_{j}=0)
				\overline{P}_{0}(\mathbf{X}={\newinstance})
				\overline{P}_{0}(Y_{\setindices^{j-1}_*}=
					\hat{\bm{y}}_{\setindices^{j-1}_*},
				Y_{\setindices_\mathcal{A}^{j-1}}=\vect{y})
			}
			{P(Y_{j}=1)
				\underline{P}_{1}(\mathbf{X}={\newinstance})
				\underline{P}_{1}(Y_{\setindices^{j-1}_*}=
					\hat{\bm{y}}_{\setindices^{j-1}_*},
				Y_{\setindices_\mathcal{A}^{j-1}}=\vect{y})
			}
	\right)^{-1},
\end{equation*}
where $\hat{\bm{y}}_{\setindices^{j-1}_*}$ is the binary vector with predicted relevant and irrelevant values. So, using the fact that minimizing $\frac{1}{1+x}$ is equal to maximizing $x$, we therefore get 
\begin{equation*}
	\max_{\vect{y}\in\{0,1\}^{\left|\setindices_{\mathcal{A}}^{j-1}\right|}} 
	\frac{P(Y_{j}=0)
		\overline{P}_{0}(\mathbf{X}={\newinstance})
		\overline{P}_{0}(Y_{\setindices^{j-1}_*}=
			\hat{\bm{y}}_{\setindices^{j-1}_*})
		\overline{P}_{0}(Y_{\setindices_\mathcal{A}^{j-1}}=
			\vect{y})
	}
	{P(Y_{j}=1)
		\underline{P}_{1}(\mathbf{X}={\newinstance})
		\underline{P}_{1}(Y_{\setindices^{j-1}_*}=
			\hat{\bm{y}}_{\setindices^{j-1}_*})
		\underline{P}_{1}(Y_{\setindices_\mathcal{A}^{j-1}}=
			\vect{y})
	}.
\end{equation*}

The first three terms of the numerator (and of the denominator) of the last equation can be omitted from the optimization problem since they do not change the optimal solution, and by applying Equations~ \eqref{eq:ncclabelupperproof}, \eqref{eq:ncclabellowerproof} and \eqref{eq:nccidm}, to the last term, we get what we sought
\begin{align*}
	&\max_{\vect{y}\in\{0,1\}^{\left|\setindices_{\mathcal{A}}^{j-1}\right|}} 
	\frac{\overline{P}_{0}(Y_{\setindices_\mathcal{A}^{j-1}}=
			\vect{y})}
	{\underline{P}_{1}(Y_{\setindices_\mathcal{A}^{j-1}}=
			\vect{y})} \\
	\iff & \max_{\vect{y}\in\{0,1\}^{\left|\setindices_{\mathcal{A}}^{j-1}\right|}} 
	\frac{\prod_{k\in\setindices_\mathcal{A}^{j-1}} \overline{P}(Y_k=y_k|Y_{j}\eq0)}
	{\prod_{k\in\setindices_\mathcal{A}^{j-1}} \underline{P}(Y_k=y_k|Y_{j}\eq1)}\\
	\iff &\max_{\vect{y}\in\{0,1\}^{\left|\setindices_{\mathcal{A}}^{j-1}\right|}} 
		\frac{\prod_{k\in\setindices_\mathcal{A}^{j-1}}\frac{n(y_k|y_j=0)+s}{n(y_j=0)+s}}
		{\prod_{k\in\setindices_\mathcal{A}^{j-1}}\frac{n(y_k|y_j=1)}{n(y_j=1)+s}}\\
	\iff &\max_{\vect{y}\in\{0,1\}^{\left|\setindices_{\mathcal{A}}^{j-1}\right|}} 
	\left[\frac{n(y_j=1)+s}{n(y_j=0)+s}
		\right]^{\left|\setindices_{\mathcal{A}}^{j-1}\right|}
	\prod_{y_i\in\vect{y}}^{~~i\in\setindices_{\mathcal{A}}^{j-1}}
		\frac{n(y_i|y_j=0)+s}{n(y_i|y_j=1)},
\end{align*}
in which it is easy see that: (1) the term $[\cdots]^{\left|\setindices_{\mathcal{A}}^{j-1}\right|}$ can be omitted, and hence, we can get $\hat{\underline{\vect{y}}}_{\setindices_{\mathcal{A}}^{j-1}}$, and (2) using similar arguments as above we can easily get the labels $\hat{\overline{\vect{y}}}_{\setindices_{\mathcal{A}}^{j-1}}$ which maximise the upper probability $\overline{P}_{\newinstance}(Y_j=1)$ of \eqref{eq:impbranching}.
\end{proof}
\mbox{}\\
\begin{proof}[\textbf{of Proposition~\ref{prop:ibcomplexity}}]
	This proof can be performed using a dichotomy algorithm (equivalent to a binary search tree), starting with $y_{k}$ last abstained label (i.e. $k=|\setindices_{\mathcal{A}}|-1$) and calculating the values $\frac{n(y_k=1|\cdot)+s}{n(y_k=1|\cdot)}$ and $\frac{n(y_k=0|\cdot)+s}{n(y_k=0|\cdot)}$, then we retain the maximal value of these last two terms (or the minimal value, whichever applies) and we go forward with second-to-last label $y_{k-1}$, but this time multiplied by the last term retained, and so on. After having obtained the lower binary path $\hat{\underline{\vect{y}}}_{\setindices_{\mathcal{A}}^{j-1}}$(or the upper binary path $\hat{\overline{\vect{y}}}_{\setindices_{\mathcal{A}}^{j-1}}$), we can directly calculate the values $\lowercondprobchain(Y_j=1)$ and $\uppercondprobchain(Y_j=1)$ (and by duality of the lower and upper probability bounds $[\lowercondprobchain(Y_j=0), \uppercondprobchain(Y_j=0)]$).
\end{proof}
\mbox{}\\
\begin{proof}[\textbf{of Proposition~\ref{prop:globalcomplexity}}]
The proof for the best-case is straighforward, because if there is not any abstained labels, the time complexity is the same than precise chaining $\mathcal{O}(m)$. The worst-case complexity, in which all inferred labels are abstained, is also easy to calculate: the first label performs a single operation, i.e. $\mathcal{O}(1)$, then the second label is also inferred in a single operation due to the number of previous abstained labels being equal to $1$ (c.f. Proposition~\ref{prop:ibcomplexity}), then the third label takes into account two previous abstained labels and performs two operations (c.f. Proposition~\ref{prop:ibcomplexity}), and the fourth label performs three operations, and so on. We therefore obtain $\mathcal{O}(\nicefrac{m(m-1)}{2}+1)$ operations which is equal to $\mathcal{O}(m^2)$ asymptotically.
\end{proof}
\newpage
\section{Supplementary results}

\begin{figure}[!th]
	\centering
	\subfigure[\sc Emotions]{\hspace{-2mm}
	   \includegraphics[width=0.33\linewidth]{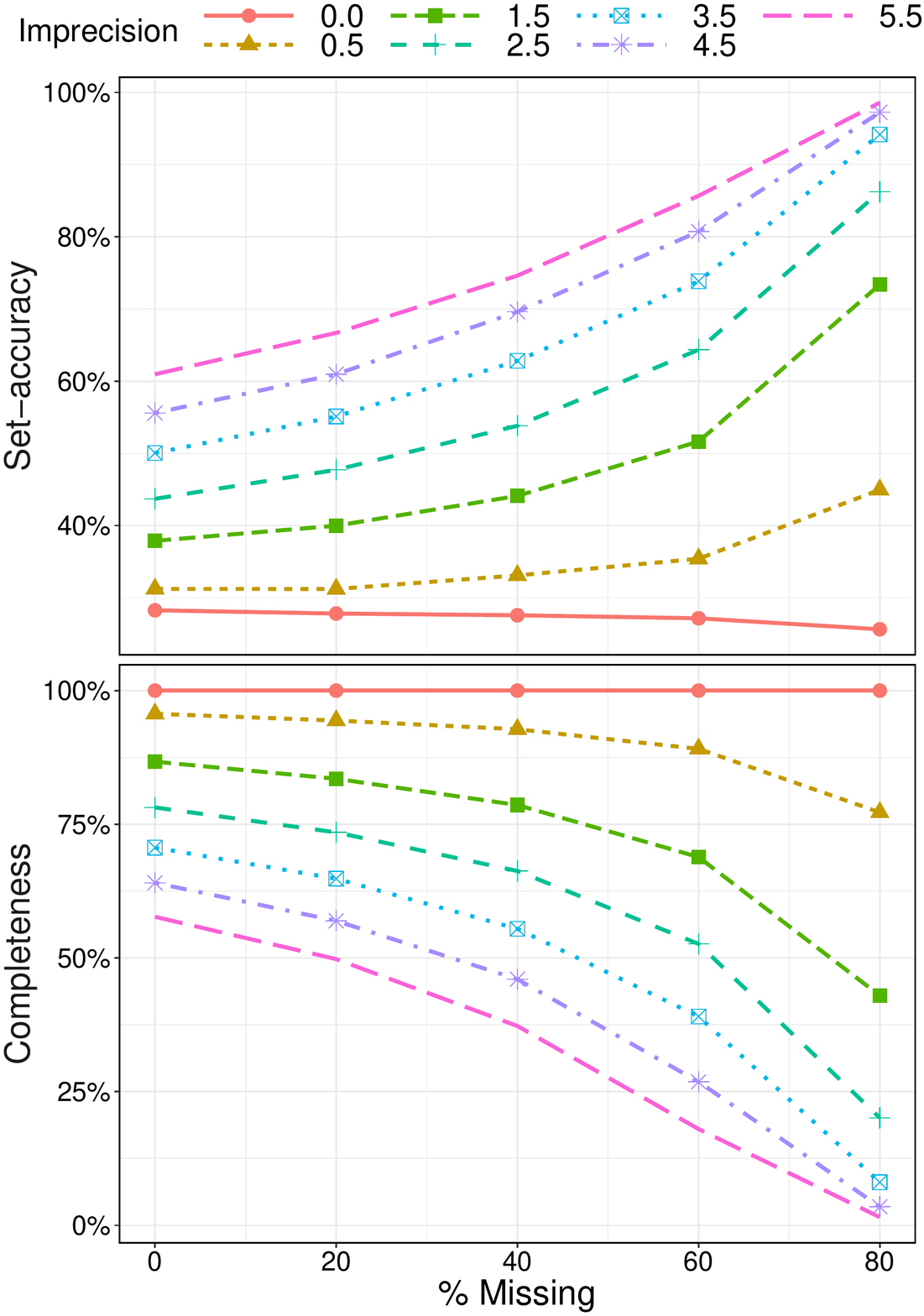}
	}%
	\subfigure[\sc Enron]{\hspace{-1mm}
		\includegraphics[width=0.33\linewidth]{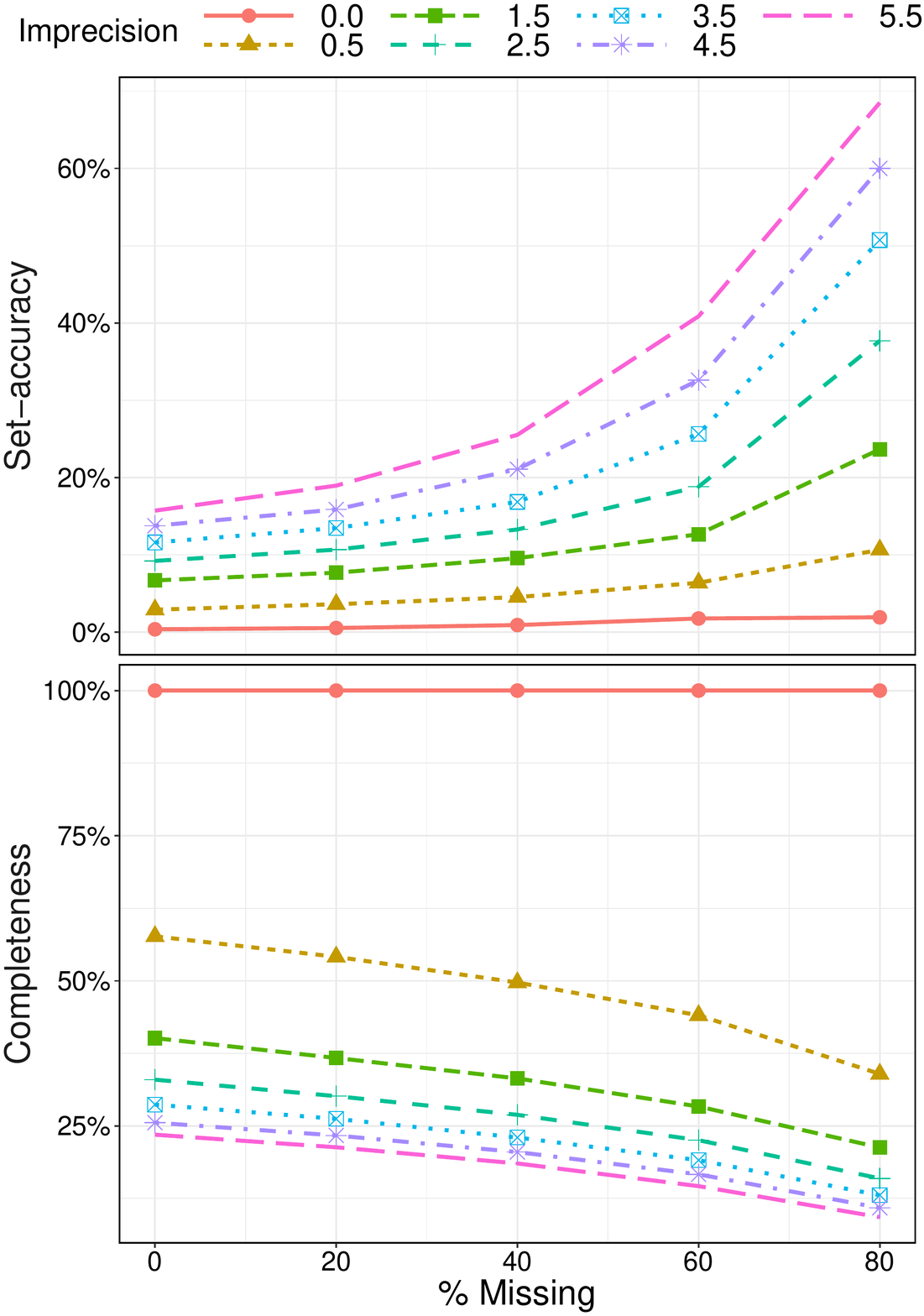}
	}%
	\subfigure[\sc Yeast]{\hspace{-1mm}
		\includegraphics[width=0.33\linewidth]{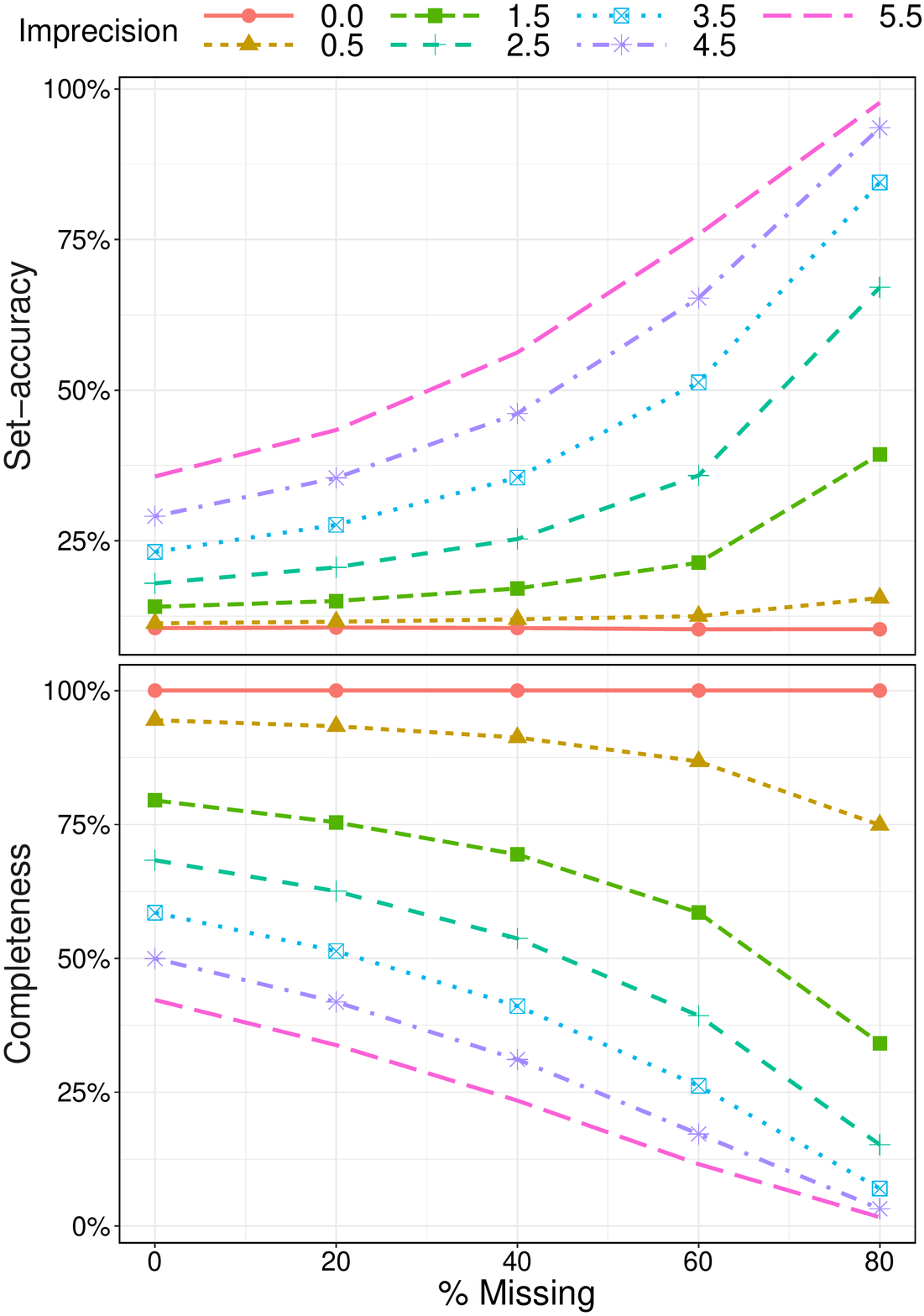}
	}\quad%
	\subfigure[\sc Scene]{\hspace{-2mm}
	   \includegraphics[width=0.33\linewidth]{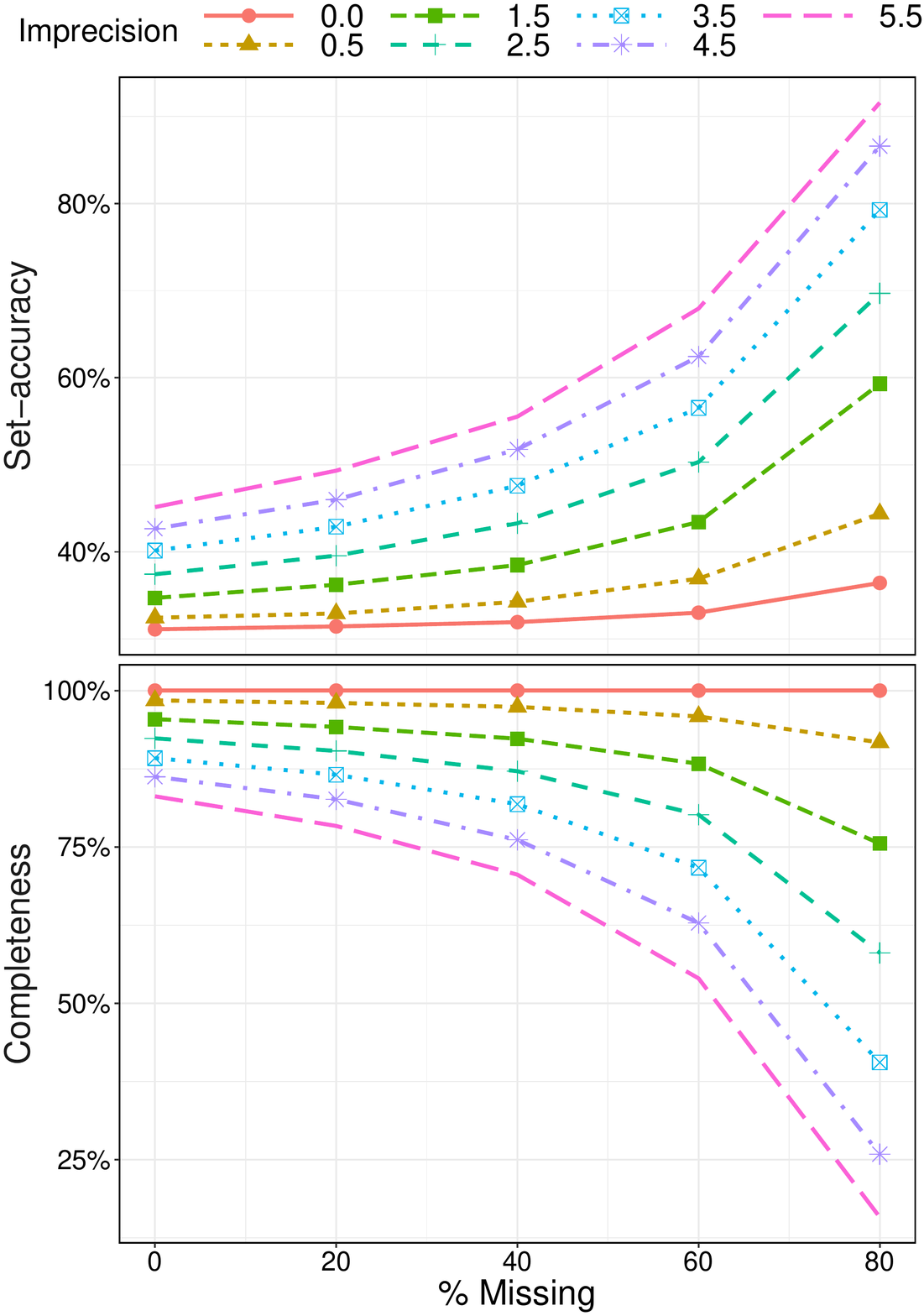}
	}%
	\subfigure[\sc Medical]{\hspace{-1mm}
		\includegraphics[width=0.33\linewidth]{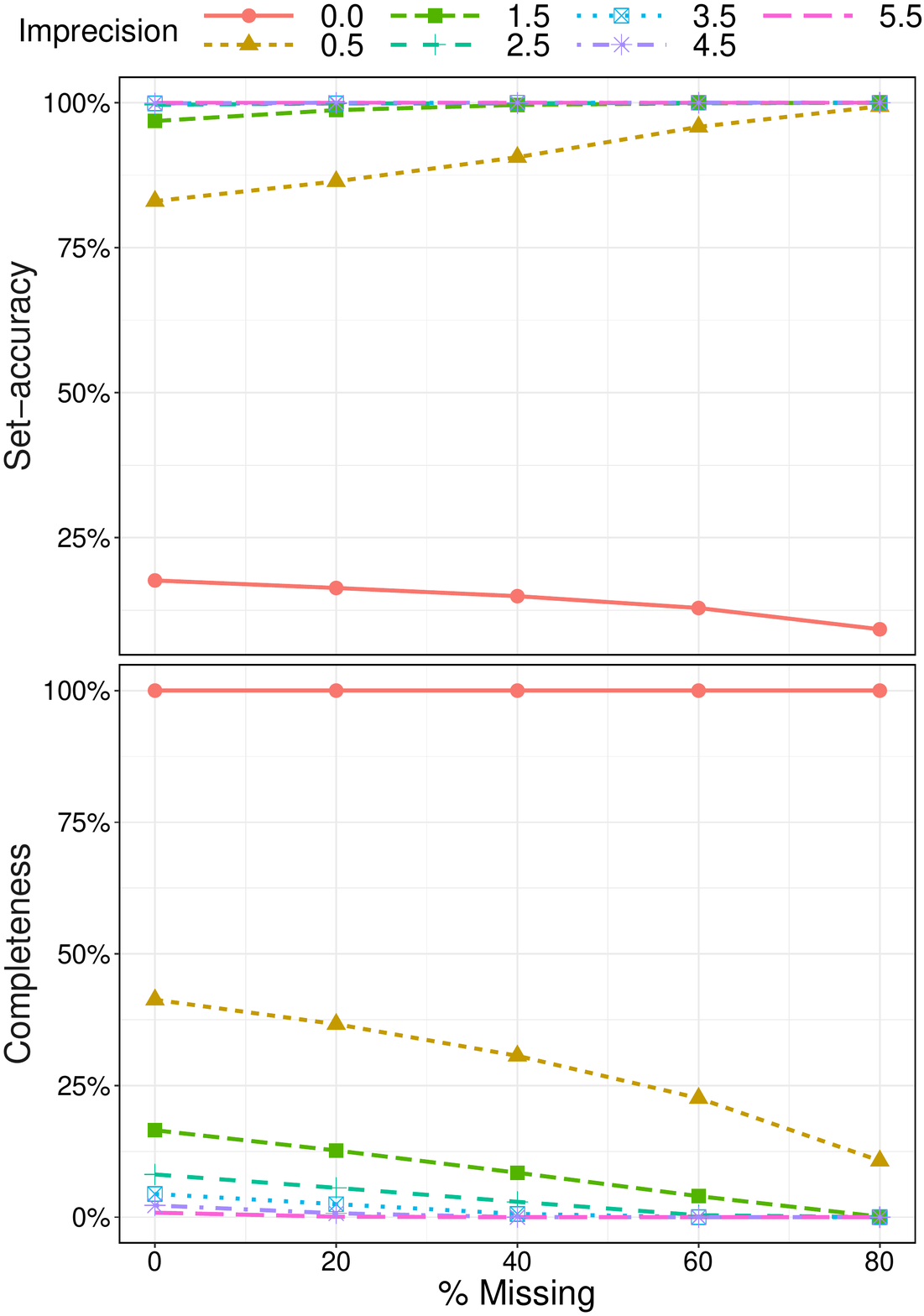}
	}%
	\subfigure[\sc CAL500]{\hspace{-1mm}
		\includegraphics[width=0.33\linewidth]{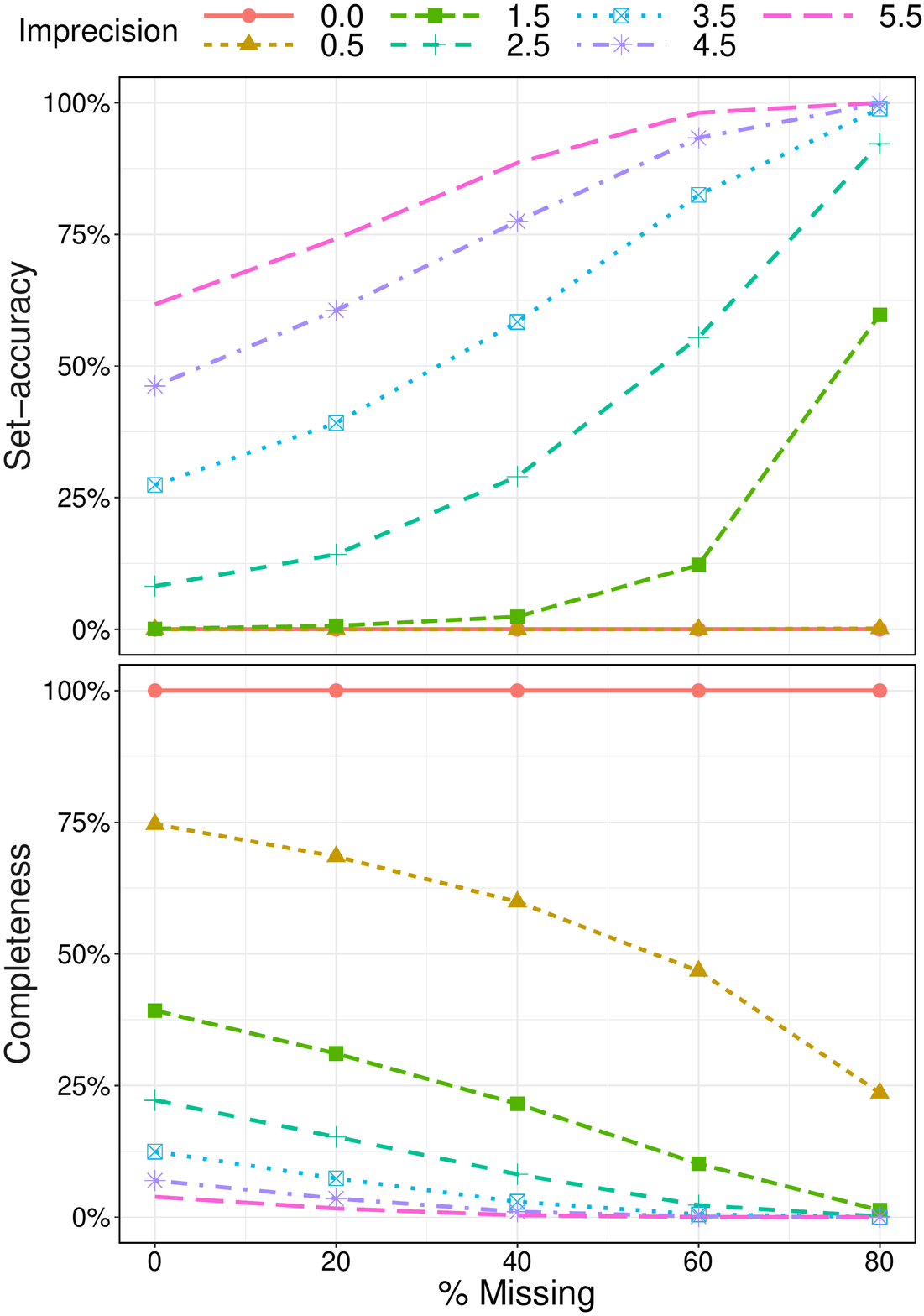}
	}%
	\caption{\textbf{Missing labels - Marginalization} Evolution of the average (\%) set-accuracy (top) and completeness (down) for each level of imprecision (a curve for each one) and discretization $z=6$, with respect the percentage of missing labels.}
\end{figure}

\begin{figure}[!th]
	\vspace{-4mm}\centering
	\subfigure[\sc Scene]{\hspace{-2mm}
	   \includegraphics[width=0.33\linewidth]{images/missing/emotions_ib_disc6}
	}%
	\subfigure[\sc Medical]{\hspace{-1mm}
		\includegraphics[width=0.33\linewidth]{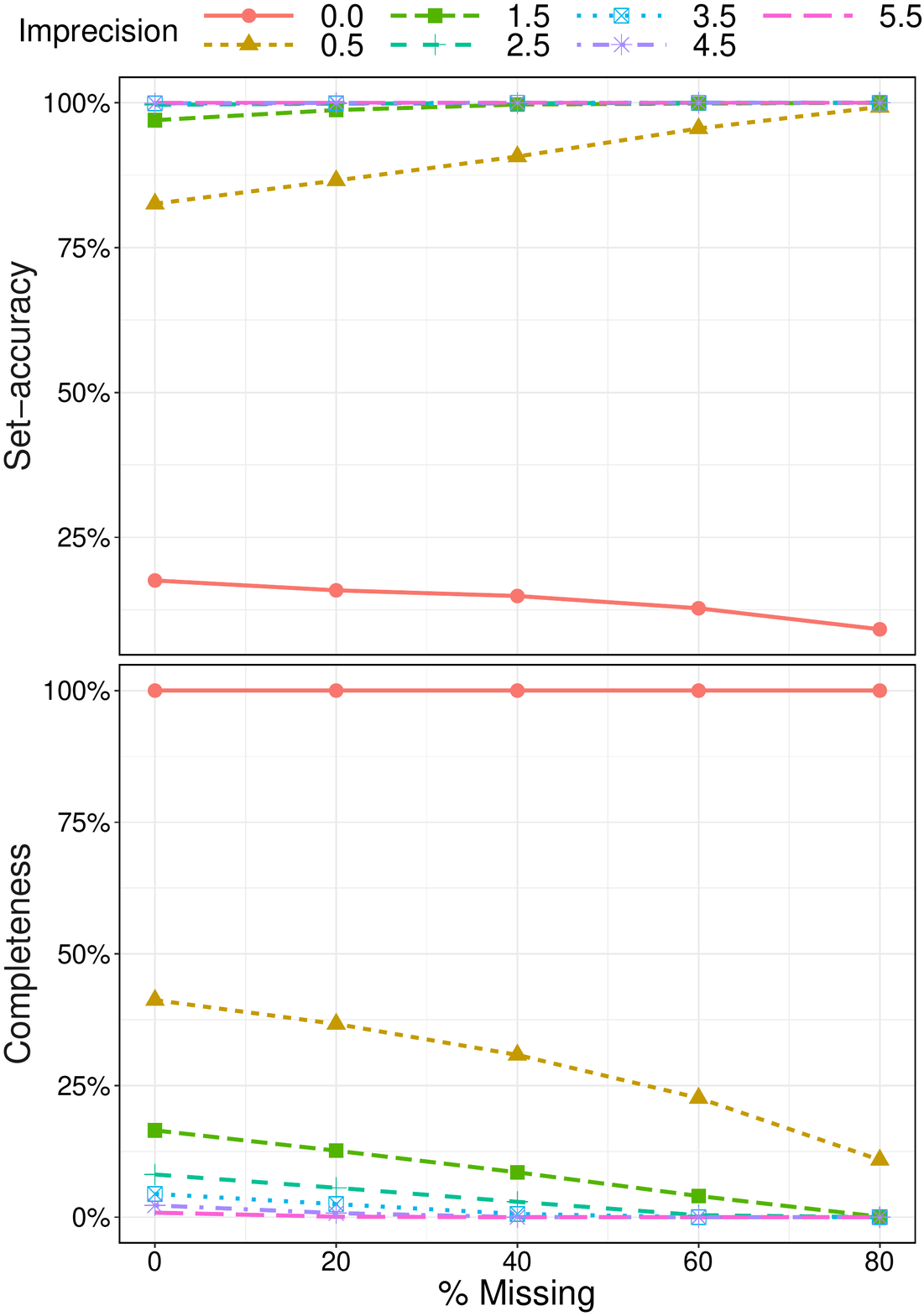}
	}%
	\subfigure[\sc CAL500]{\hspace{-1mm}
		\includegraphics[width=0.33\linewidth]{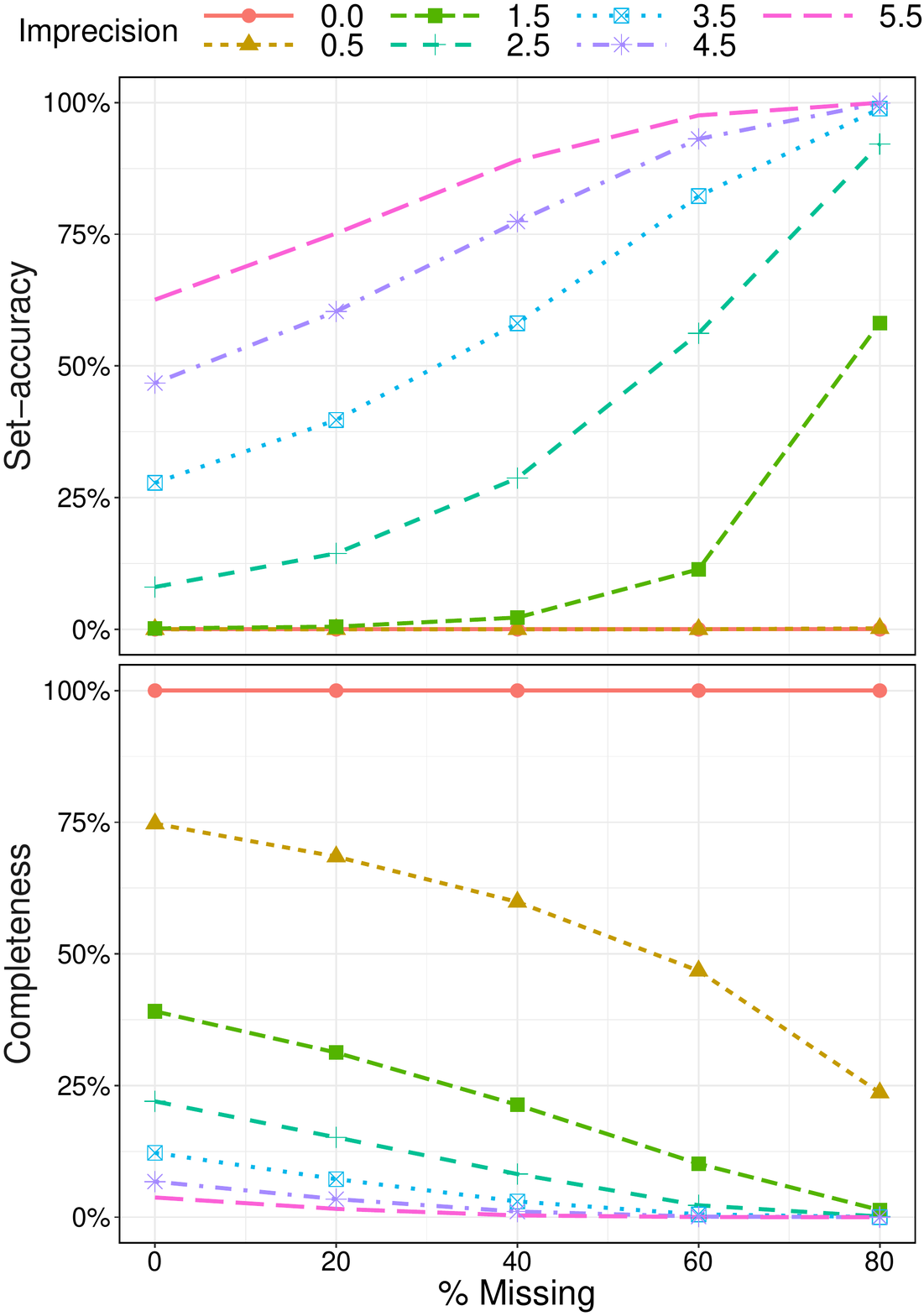}
	}%
	\vspace{-2mm}\caption{\textbf{Missing labels - Imprecise Branching} Evolution of the average (\%) set-accuracy (top) and completeness (down) for each level of imprecision (a curve for each one) and a discretization $z=6$, with respect to the percentage of missing labels.}
\end{figure}